\documentclass[letterpaper]{article}
\usepackage{aaai}
\usepackage{times}
\usepackage{helvet}
\usepackage{courier}
\usepackage{booktabs}
\usepackage{multirow}
\IfFileExists{adjustbox.sty}{\usepackage{adjustbox}}{}
\usepackage{microtype}
\usepackage{amsmath}
\usepackage{amssymb}
\usepackage{graphicx}
\usepackage{pdfpages}
\makeatletter
\@ifpackageloaded{adjustbox}{}{%
\newsavebox{\adjustboxfallbackbox}
\newenvironment{adjustbox}[1]{\begin{lrbox}{\adjustboxfallbackbox}}{\end{lrbox}\resizebox{\textwidth}{!}{\usebox{\adjustboxfallbackbox}}}
}
\makeatother
\usepackage{float}
\usepackage{tikz}
\IfFileExists{fontawesome5.sty}{\usepackage{fontawesome5}}{}
\usetikzlibrary{arrows.meta,positioning,fit,backgrounds,calc,shapes.geometric}
\definecolor{panelA}{HTML}{EAF2EA}
\definecolor{panelB}{HTML}{EAF0FA}
\definecolor{boxedge}{HTML}{6B7280}
\definecolor{accentA}{HTML}{2E7D5B}
\definecolor{accentB}{HTML}{2F5BAA}
\definecolor{subcap}{HTML}{4B5563}
\usepackage{colortbl}
\usepackage{placeins}
\usepackage{afterpage}
\usepackage[compact]{titlesec}
\usepackage{enumitem}
\titlespacing*{\section}{0pt}{1.2ex plus 0.3ex minus 0.2ex}{0.6ex}
\titlespacing*{\subsection}{0pt}{1.0ex plus 0.2ex minus 0.2ex}{0.4ex}
\titlespacing*{\paragraph}{0pt}{0.6ex plus 0.2ex minus 0.1ex}{0.6em}
\setlist[itemize]{nosep,topsep=2pt,partopsep=0pt}
\definecolor{bestgreen}{HTML}{C8E6C9}   
\IfFileExists{stfloats.sty}{\usepackage{stfloats}}{}  
\usepackage{hyperref}
\hypersetup{
  colorlinks=true,
  linkcolor=[rgb]{0.10,0.20,0.75},
  citecolor=[rgb]{0.10,0.20,0.75},
  urlcolor=[rgb]{0.05,0.35,0.85},
  pdftitle={bioMoR: Biology-Guided Mixture-of-Recursions for Effective Genomic Learning},
  pdfauthor={Koushik Howlader, Tirtho Roy, Md Tauhidul Islam, Wei Le},
  pdfkeywords={single-cell genomics, multi-omics, transformers, parameter efficiency, marker genes, recursive computation, biological knowledge}
}
\makeatletter
\def\@lbibitem[#1]#2{\item\if@filesw
{ \def\protect##1{\string ##1\space}\immediate
\write\@auxout{\string\bibcite{#2}{#1}}}\fi
\@ifundefined{hyper@anchorstart}{}{\Hy@raisedlink{\hyper@anchorstart{cite.#2}\hyper@anchorend}}%
\ignorespaces}
\makeatother

\title{bioMoR: Biology-Guided Mixture-of-Recursions for Effective Genomic Learning}
\author{Koushik Howlader$^{1}$, Tirtho Roy$^{1}$,
Md Tauhidul Islam$^{2}$, Wei Le$^{1}$\\
$^{1}$Iowa State University, Iowa, USA\\
$^{2}$Stanford University, CA, USA}

\begin{document}
\maketitle

\begin{abstract}

Transformer models for high-dimensional omics analysis process thousands of genes or pathways, although only a subset requires deep computation. Mixture-of-Recursions (MoR) improves efficiency through adaptive token-choice or expert-choice routing. We propose bioMoR, which, to the best of our knowledge, is the first framework to apply MoR to gene-level and pathway-level learning. Our contributions include identifying three locations for integrating structured biological knowledge within an MoR backbone: graph-based information sharing refines token embeddings, a structural bias guides self-attention toward biologically related tokens, and a graph-aware router uses neighborhood information to determine each token's recursion depth. These techniques are centered on our insight that additional knowledge of token interaction can effectively help models construct embeddings and select which tokens should be learned more deeply. Across eight benchmarks spanning diverse omics data types and evaluated under a unified five-fold cross-validation protocol, bioMoR improves average macro-F1 by $8.2$ percentage points and balanced accuracy by $7.1$ percentage points over the strongest biology-agnostic MoR baseline while using $75$ percent fewer parameters and up to $58$ percent fewer FLOPs than a non-recursive Transformer. The selected marker genes or pathways provide biological interpretability, while their token-specific recursion depths reveal how computation is allocated.

\end{abstract}

\begin{figure*}[t]
\centering
\includegraphics[width=0.80\linewidth]{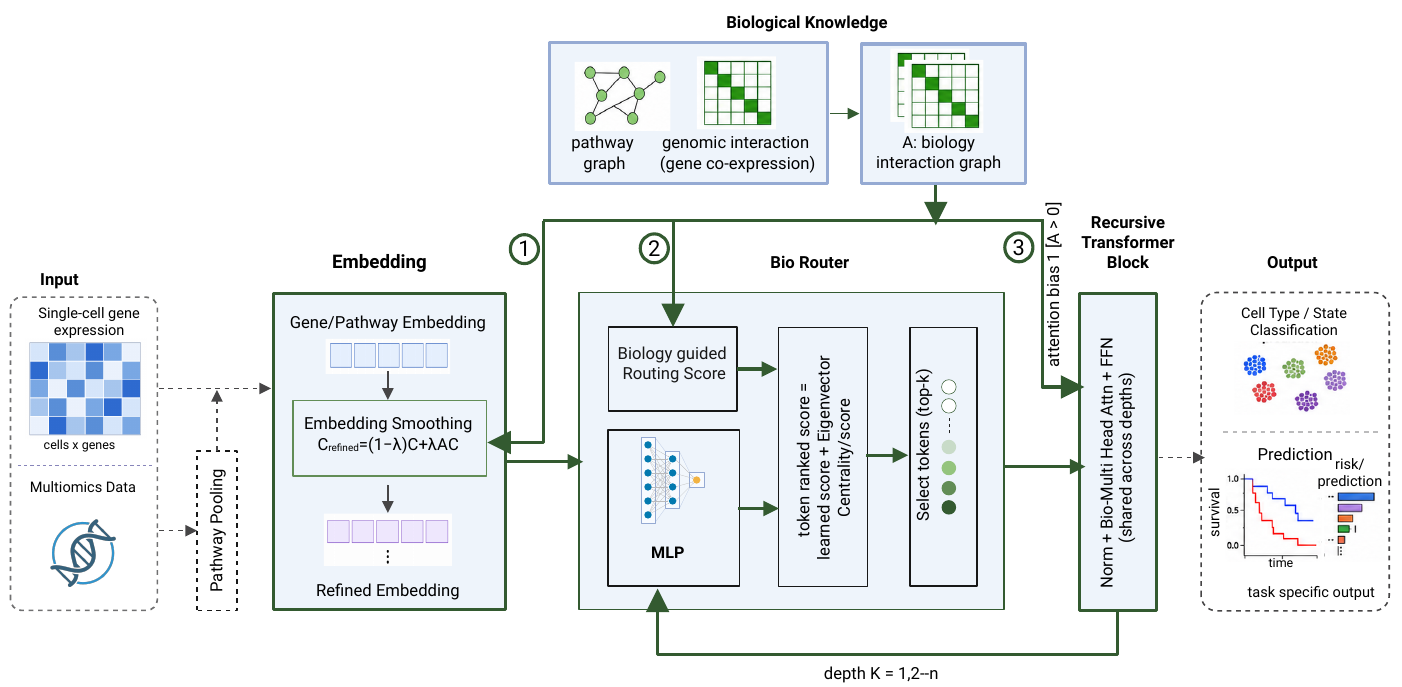}
\caption{\footnotesize\textbf{Overview of bioMoR.} Single-cell and multi-omics inputs are represented
as gene or pathway tokens. Fixed biological knowledge is incorporated at three sites:
\textbf{(1)} embedding smoothing refines related tokens, \textbf{(2)} biology-guided routing selects
tokens for further computation, and \textbf{(3)} attention bias promotes interactions between
biologically related tokens in the weight-shared recursive Transformer block.}
\label{fig:overview}
\end{figure*}

\section{Introduction}

High-throughput sequencing now measures thousands of genes, pathways, or molecular features per
sample. Transformers are attractive for these data because attention can model long-range molecular
dependencies; recent single-cell foundation models, including
scBERT~\cite{scbert2022}, scGPT~\cite{cui2024scgpt},
scFoundation~\cite{hao2024large}, and Geneformer~\cite{theodoris2023transfer}, show strong results
for annotation and representation learning. However, the standard Transformers spend the same deep computation on every token, even though genomic
signal is usually sparse: a cell type, tumor state, or metastatic phenotype is often driven by a
small set of marker genes, regulatory programs, or pathways. For computational efficiency and accuracy, a useful genomic model should not waste computation on weakly informative features while under-refining the molecular
programs that explain the phenotype. 
The effective modeling thus should answer two questions at
the same time: which genes or pathways should be kept, and which of those tokens should receive more
model computation?


\begin{center}
\setlength{\fboxsep}{4pt}
\fbox{\begin{minipage}{0.88\columnwidth}
\footnotesize
\textbf{Guiding assumption.} 
We should perform computation and learning in a selective and structured manner, based on the semantics of omics data. The model should retain biologically meaningful marker gene/pathway tokens and spend more recursive
computation on the tokens whose biological neighborhoods are most predictive for the task.
\end{minipage}}
\end{center}

Existing methods address only part of this assumption. Efficient attention methods such as
Linformer~\cite{wang2020linformer}, Performer~\cite{choromanski2021rethinking}, and
Nystr\"omformer~\cite{xiong2021nystromformer} reduce sequence cost through low-rank or kernel
approximations. Adaptive computation methods such as ACT~\cite{graves2016adaptive},
mixture-of-experts~\cite{shazeer2017outrageously},
mixture-of-depths~\cite{raposo2024mixture}, and
mixture-of-recursions~\cite{bae2025mixture} route tokens dynamically. However, these
approaches are biology-agnostic, for example, they do not know whether  two genes express together, share a pathway, or contribute to a same biological process. Conversely, biology-aware resources and models such as
CellMarker~\cite{hu2023cellmarker}, PanglaoDB~\cite{franzen2019panglaodb},
Reactome~\cite{gillespie2022reactome}, genomic interaction graphs~\cite{islam2023cartography},
Pathformer~\cite{pathformer2023}, and scBiGNN~\cite{scbignn2023} provide and consider useful
semantic structures of omics data,
however, they are usually used for feature engineering, architectural constraints, or post-hoc interpretation rather than for selecting tokens and deciding which tokens should receive deeper computation.

We propose \textbf{bioMoR}, a biology-guided Mixture-of-Recursions framework for genomic learning. To the best of our knowledge, it is the first work that integrates biology knowledge and Mixture-of-Recursion architecture to drive gene or pathway token selections for adaptive learning.

As shown in Figure~\ref{fig:overview}, bioMoR first compresses high-dimensional inputs into marker-gene or Reactome pathway tokens. It then
uses a shared recursive Transformer block and routes each
token to an adaptive number of recursive steps. The key change from biology-agnostic MoR~\cite{bae2025mixture} is that we inject relevant biology knowledge into the model at three sites: it smooths token embeddings, biases attention
toward related tokens, and guides the MoR router to decide token-specific recursion depth. Thus, 
biological structure and semantics becomes a control signal for adaptive computation, not only a static prior.

This leads to four questions in our evaluation: does bioMoR improve predictive performance (RQ1), which bio-knowledge injection
site matters (RQ2), whether it indeed makes computation more efficient (RQ3), and which routing, marker
budget, and model width settings are optimal (RQ4). 

Across eight diverse genomic learning datasets, bioMoR improves average macro-F1 by $8.2$ points and balanced accuracy by $7.1$ points, while using $75\%$ fewer parameters and up to $58\%$ fewer FLOPs than traditional Transformers. It also achieved $0.92$ linear-probe accuracy and successfully selected biologically relevant pathways for deeper computation, confirmed by prior literature.

In summary,  our work makes the following contributions:
\begin{itemize}
\item We introduce bioMoR, the first Mixture-of-Recursions (MoR) framework for gene- and pathway-level
omics learning.
\item We integrated biology knowledge, including gene co-expression and pathway-pathway relations, to improve representation learning, attention learning,
and adaptive routing leveraging MoR backbone.
\item We conducted extensive evaluation and show that our techniques significantly improves F1 and balanced accuracy and reduce model parameters as well as the FLOP cost, and they are generally applicable across single-cell and multi-omics benchmarks.
\end{itemize}

\section{Related Work}
\label{sec:related_work}

\paragraph{Genomic representation learning.}
Transformer models such as scBERT~\cite{scbert2022}, scGPT~\cite{cui2024scgpt}, and
Geneformer~\cite{theodoris2023transfer} learn gene representations from large single-cell datasets
and support tasks such as cell-type annotation. These models are effective, but normally apply
the same model depth to every input token.

\paragraph{Biological knowledge in omics models.}
P-NET uses known biological hierarchies for cancer prediction \cite{elmarakeby2021biologically},
while Pathformer~\cite{pathformer2023} and scBiGNN~\cite{scbignn2023} use pathway or
gene-interaction information to guide genomic learning. These methods show the value of biological knowledge, but they do
not use it to decide how much computation each token should receive.

\paragraph{Adaptive computation.}
Adaptive Computation Time~\cite{graves2016adaptive},
mixture-of-experts~\cite{shazeer2017outrageously},
mixture-of-depths~\cite{raposo2024mixture}, and MoR~\cite{bae2025mixture} allow different tokens to
receive different amounts of computation. However, their
routing decisions are biology-agnostic. bioMoR connects these two research directions by using
biological knowledge to guide token representations, attention, and recursion depth.

\section{Method}
\label{sec:method}

First, in \emph{Genomic Tokenization} and \emph{Mixture-of-Recursions},  we provide background on two existing components \cite{cui2024scgpt,elmarakeby2021biologically,bae2025mixture}, on top of which we built bioMoR. Second, in \emph{bioMoR}, we explained our novel techniques of integrating biological knowledge to guide MoR learning. See Figure~\ref{fig:overview} for our overall workflow. 

\subsection{Genomic Tokenization}

Let $x\in\mathbb{R}^{N\times C}$ denote one genomic sample with $N$ genes and $C$ molecular
channels. Single-cell gene expression measures transcript abundance and has $C{=}1$
\cite{wang2009rnaseq}. Multi-omics inputs may additionally contain mutation indicators, which
record DNA-sequence alterations \cite{vogelstein2013cancer}, and copy-number variation (CNV), which
records gains or losses of DNA segments \cite{feuk2006structural}. 
$T=[c_1,\ldots,c_M]\in\mathbb{R}^{M\times d}$. 

\paragraph{Single-cell marker tokens.}
A single cell has thousands of genes, but only a few are informative. We therefore turn the $N$ genes
into $M\ll N$ marker tokens. This makes the later attention cheap ($\mathcal{O}(M^2 d)$) and lets each
token be read as one named marker gene. First, each gene $i$ is embedded as an identity vector plus
its expression value,
\begin{equation}
    t_i = e_i + W_v x_i,\qquad e_i\in\mathbb{R}^{d},\ W_v\in\mathbb{R}^{d\times C},
\end{equation}
following the gene-plus-value scheme of single-cell foundation models
\cite{cui2024scgpt,theodoris2023transfer}. We then keep $M$ learnable queries; query $m$ softly
picks genes through keys $k_i=W_k e_i$,
\begin{equation}
    w_{mi} =
    \frac{\exp(q_m^\top k_i /(\tau\sqrt{d}))}
    {\sum_{j=1}^{N}\exp(q_m^\top k_j /(\tau\sqrt{d}))},
\end{equation}
and sums them into a token $c_m = \sum_{i=1}^{N} w_{mi} t_i$. At test time, query $m$ collapses to its
top gene $g_m=\arg\max_i w_{mi}$, so every token names a marker gene. 


\paragraph{Pathway tokens.}
For multi-omics cancer cohorts we instead group genes into Reactome pathways, which represent
curated sets of genes participating in the same biological process \cite{gillespie2022reactome}.
Each token is one pathway, built by pooling its member genes---mean pooling for dense expression and
copy-number channels, and sum/burden pooling for sparse mutation channels. Every token is thus a
named pathway.

\subsection{Mixture-of-Recursions}

Let $T=[c_1,\ldots,c_M]$ be the marker/pathway token matrix. MoR applies one shared Transformer
block $f_\theta$ up to $K$ times:
\begin{equation}
    H^{(0)} = T,\qquad H^{(t+1)} = f_\theta(H^{(t)}),\quad t=0,\ldots,K-1 .
\end{equation}
Sharing makes parameter count independent of $K$, unlike $K$ independent Transformer blocks
\cite{dehghani2019universal,lan2020albert,bae2025mixture}. A router decides which tokens continue.
Expert-choice keeps the top-$\lceil c_tM\rceil$ tokens at step $t$; token-choice lets each token
self-select a depth with load balancing \cite{shazeer2017outrageously}. The number of active steps,
\begin{equation}
    d_m = \sum_{t=1}^{K} \mathbf{1}\{m\ \text{is kept at step}\ t\},
\end{equation}
is the token-specific recursion depth. 


\subsection{bioMoR}

\begin{center}
\setlength{\fboxsep}{4pt}
\fbox{\begin{minipage}{0.92\columnwidth}
\footnotesize
\textbf{Algorithm 1: bioMoR forward pass.}
\begin{tabbing}
\hspace{1.2em}\=\hspace{1.2em}\=\kill
\textbf{Input:} $x$, biological knowledge $B$\\
\> token budget $M$, max depth $K$\\
$T \leftarrow$ marker-gene or pathway tokens from $x$\\
$A \leftarrow$ row-normalized fixed knowledge matrix from $B$\\
$H \leftarrow (1-\lambda)T+\lambda A T$;\quad $\mathcal{A}\leftarrow\{1,\ldots,M\}$\\
\textbf{for} $t=1,\ldots,K$ \textbf{while} $\mathcal{A}\ne\emptyset$ \textbf{do}\\
\> $H_{\mathcal{A}} \leftarrow f_\theta(H_{\mathcal{A}};A)$ \quad \textit{reuse the same block}\\
\> $r_{\mathcal{A}} \leftarrow$ biology-aware router scores\\
\> $\mathcal{A}_{\mathrm{next}} \leftarrow$ tokens routed to the next step\\
\> tokens in $\mathcal{A}\setminus\mathcal{A}_{\mathrm{next}}$ stop and keep current states\\
\> $\mathcal{A}\leftarrow\mathcal{A}_{\mathrm{next}}$ \quad \textit{return kept tokens to recursion}\\
\textbf{end for}\\
$\hat y \leftarrow$ classifier$\big(\mathrm{pool}(H)\big)$\\
\textbf{Train:} minimize class-weighted CE plus auxiliary losses\\
\textbf{Output:} $\hat y$, markers/pathways, recursion depths $d_m$
\end{tabbing}
\end{minipage}}
\end{center}

\paragraph{Biological-knowledge construction.}
Following Genomap \cite{islam2023cartography}, we construct a gene {\it co-expression} matrix that
summarizes pairwise gene relationships. Its entries are partial correlations computed from the
inverse covariance of the gene-expression matrix and restricted to the selected markers; this
matrix forms the single-cell biological knowledge $B$. For multi-omics data, we use curated Reactome pathway
relationships and treat them as undirected: $B_{ij}=B_{ji}=1$ when pathways $i$ and $j$ are related,
and both entries are zero otherwise \cite{gillespie2022reactome}.

Let $D$ be the diagonal matrix of row sums of $B$. We use the fixed, row-normalized matrix
\begin{equation}
    A = D^{-1}B .
\end{equation}
The same biological knowledge is used for embedding smoothing (Site 1), attention bias
(Site 2), and depth routing (Site 3).

\paragraph{Site 1: biology-guided embedding smoothing.}
Before recursion, we smooth the tokens using the biological-knowledge graph $A$, so that information of biologically
related tokens is also integrated into the embedding of the token:
\begin{equation}
    \widetilde T = (1-\lambda)T + \lambda A T,
\end{equation}
where $\lambda$ is learned. This denoises each marker/pathway token toward its biological neighbours
before the repeated recursive updates.

\paragraph{Site 2: attention bias.}
Inside the recursive block, the \emph{same} biological-knowledge graph $A$ biases self-attention:
\begin{equation}
    \mathrm{Attn}(Q,K,V)
    =
    \mathrm{softmax}\!\left(
        \frac{QK^\top}{\sqrt d}
        +
        \lambda_{\mathrm{attn}}\mathbf{1}[A>0]
    \right)V .
\end{equation}
This encourages biologically related tokens to exchange information during refinement.

\paragraph{Site 3: biology-aware routing.}
  The original MoR router uses the current token representation $h_m^{(t)}$ to decide whether token
  $m$ should continue to the next recursion step. bioMoR keeps this original score and adds a learned
  correction based on the token's biological neighborhood:
  \begin{equation}
      \widetilde r_m^{(t)}
      =
      \frac{w_r^\top h_m^{(t)}}{\tau_r}
      +
      \sigma(\gamma_t)
      \phi_\psi\!\left(
          [\,A H^{(t)},\ H^{(t)}-A H^{(t)}\,]_m
      \right).
  \end{equation}
  The first term is the original MoR routing score. In the second term, $AH^{(t)}$ summarizes
  information from biologically related tokens, while $H^{(t)}-AH^{(t)}$ shows how the token differs
  from its biological neighborhood. The small network $\phi_\psi$ uses these two signals to adjust the
  routing score, and $\sigma(\gamma_t)$ controls how much correction is added at recursion step $t$.
  Because $\phi_\psi$ is initialized to output zero, the correction is zero at the start of training
  and bioMoR initially behaves like biology-agnostic MoR. During training, the correction becomes
  nonzero when biological information helps reduce the task loss; otherwise, it can remain close to
  zero. This design prevents noisy biological knowledge from affecting routing at the beginning while
  allowing the model to use it when helpful.
\paragraph{Training and efficiency.}
The main task loss is class-weighted cross-entropy. During training we minimize
\begin{equation}
    \mathcal L
    =
    \mathcal L_{\mathrm{CE}}
    + \alpha\mathcal L_{\mathrm{suf}}
    + \beta\mathcal L_{\mathrm{div}}
    + \gamma\mathcal L_{\mathrm{cmp}}
    + \eta\mathcal L_{\mathrm{router}},
\end{equation}
where the auxiliary terms encourage marker sufficiency ($\mathcal L_{\mathrm{suf}}$), marker
diversity ($\mathcal L_{\mathrm{div}}$), marker compression/sparsity ($\mathcal L_{\mathrm{cmp}}$),
and stable router logits and balanced token-choice routing ($\mathcal L_{\mathrm{router}}$).
In expert-choice routing, each recursion step selects a fixed number of the highest-scoring tokens.
This fixed capacity directly controls how many tokens are processed at each step, so an additional
load-balancing loss is not needed \cite{zhou2022expertchoice,bae2025mixture}. To quantify
computational cost, we report FLOPs, a hardware-independent count
of the arithmetic operations performed by the Transformer stack \cite{schwartz2020green}:
\begin{equation}
    \Phi_{\mathrm{eff}}=\sum_{t=1}^{K}\left(4a_t^2d+4a_tdd_{\mathrm{ff}}\right),
\end{equation}
where $a_t$ is the mean number of active tokens at recursion step $t$, $d$ is the model width, and
$d_{\mathrm{ff}}$ is the hidden dimension of the feed-forward network.


\afterpage{%
\begin{table*}[t]
\centering
\caption{Macro-F1 and balanced accuracy (mean$\pm$SD) from five-fold stratified cross-validation
on eight benchmark datasets. All rows use the same folds and operating point; Recursive, MoR, and
bioMoR use $K{=}4$. Avg.\ is the arithmetic mean across the eight displayed datasets. Underlining
marks the top two results per dataset.}
\label{tab:selected_results}
\setlength{\tabcolsep}{2.7pt}
\renewcommand{\arraystretch}{1.10}
\begin{adjustbox}{max width=\textwidth}
\begin{tabular}{@{}c l l c c c c c c c c c c@{}}
\multicolumn{4}{c}{} &
\multicolumn{5}{c}{\textbf{Single-cell}} &
\multicolumn{3}{c}{\textbf{Pathway-based}} & \multicolumn{1}{c}{} \\
\cmidrule(lr){5-9}\cmidrule(lr){10-12}
\textbf{Metric} & \textbf{Model} & \textbf{Routing} & \textbf{Param} &
\textbf{Segerstolpe} & \textbf{Lung} & \textbf{T-cell} & \textbf{Muraro} &
\textbf{Spleen} & \textbf{BLCA} & \textbf{PAN-2M} & \textbf{PAN-3M} & \textbf{Avg.} \\
\midrule

\multirow{6}{*}{\rotatebox[origin=c]{90}{\footnotesize\textbf{Macro-F1}}}
& Vanilla & -- & 300K
& $62.5 \pm 11.2$ & $72.5 \pm 2.5$ & $52.1 \pm 3.5$ & $80.2 \pm 4.3$
& $48.3 \pm 3.7$ & $40.6 \pm 5.8$ & $75.6 \pm 3.5$ & $95.2 \pm 1.5$ & $65.9$ \\
& Recursive & -- & 75K
& $61.4 \pm 7.3$ & $72.9 \pm 2.9$ & $50.5 \pm 3.5$ & $75.8 \pm 4.1$
& $49.4 \pm 3.4$ & $40.2 \pm 1.1$ & $72.3 \pm 1.8$ & $94.3 \pm 1.1$ & $64.6$ \\
\addlinespace[2pt]
& MoR & Expert & 75K
& $64.3 \pm 11.6$ & $72.3 \pm 2.7$ & $50.9 \pm 4.7$ & $74.8 \pm 4.7$
& $53.2 \pm 1.3$ & $\underline{43.7 \pm 2.8}$ & $75.5 \pm 3.0$ & $94.3 \pm 1.3$ & $66.1$ \\
& & Token & 75K
& $60.7 \pm 5.8$ & $73.3 \pm 2.4$ & $50.4 \pm 2.2$ & $81.6 \pm 2.7$
& $51.1 \pm 3.1$ & $40.6 \pm 6.9$ & $78.0 \pm 4.6$ & $95.3 \pm 1.6$ & $66.4$ \\
\addlinespace[2pt]
& \textbf{bioMoR (ours)} & Expert & 75K
& $\underline{72.2 \pm 4.5}$ & $\underline{80.0 \pm 1.2}$
& $\underline{70.0 \pm 1.9}$ & $\underline{83.7 \pm 4.9}$
& $\underline{59.7 \pm 0.8}$ & $\underline{47.1 \pm 7.1}$
& $\underline{85.8 \pm 1.5}$ & $\underline{98.0 \pm 0.4}$ & $\mathbf{74.6}$ \\
& & Token & 75K
& $\underline{75.8 \pm 2.8}$ & $\underline{79.7 \pm 0.9}$
& $\underline{69.8 \pm 1.2}$ & $\underline{81.9 \pm 6.8}$
& $\underline{61.0 \pm 0.6}$ & $39.5 \pm 5.9$
& $\underline{86.0 \pm 1.9}$ & $\underline{96.4 \pm 1.6}$ & $\mathbf{73.8}$ \\

\midrule

\multirow{6}{*}{\rotatebox[origin=c]{90}{\footnotesize\textbf{Balanced accuracy}}}
& Vanilla & -- & 300K
& $71.7 \pm 10.3$ & $75.2 \pm 1.5$ & $55.2 \pm 5.7$ & $79.1 \pm 3.3$
& $55.3 \pm 1.4$ & $48.3 \pm 2.8$ & $74.1 \pm 4.0$ & $94.9 \pm 1.9$ & $69.2$ \\
& Recursive & -- & 75K
& $69.6 \pm 9.1$ & $75.3 \pm 1.9$ & $56.5 \pm 4.6$ & $77.8 \pm 2.4$
& $56.9 \pm 1.5$ & $48.5 \pm 2.8$ & $74.4 \pm 1.6$ & $93.6 \pm 1.7$ & $69.1$ \\
\addlinespace[2pt]
& MoR & Expert & 75K
& $69.2 \pm 6.1$ & $76.5 \pm 1.8$ & $56.6 \pm 6.1$ & $75.5 \pm 2.3$
& $57.8 \pm 2.5$ & $\underline{49.8 \pm 5.0}$ & $74.6 \pm 3.1$ & $93.7 \pm 3.1$ & $69.2$ \\
& & Token & 75K
& $66.4 \pm 8.0$ & $76.1 \pm 2.2$ & $56.6 \pm 4.1$ & $80.6 \pm 8.5$
& $56.5 \pm 1.4$ & $46.7 \pm 7.8$ & $73.5 \pm 2.2$ & $93.7 \pm 4.6$ & $68.8$ \\
\addlinespace[2pt]
& \textbf{bioMoR (ours)} & Expert & 75K
& $\underline{78.8 \pm 4.9}$ & $\underline{80.4 \pm 0.6}$
& $\underline{70.4 \pm 1.6}$ & $\underline{88.9 \pm 4.0}$
& $\underline{64.7 \pm 1.6}$ & $\underline{48.7 \pm 7.5}$
& $\underline{82.9 \pm 3.3}$ & $\underline{95.4 \pm 1.3}$ & $\mathbf{76.3}$ \\
& & Token & 75K
& $\underline{73.3 \pm 7.0}$ & $\underline{82.0 \pm 1.5}$
& $\underline{71.3 \pm 1.4}$ & $\underline{83.4 \pm 3.7}$
& $\underline{64.1 \pm 1.1}$ & $48.3 \pm 2.2$
& $\underline{83.4 \pm 2.6}$ & $\underline{96.1 \pm 1.9}$ & $\mathbf{75.2}$ \\

\addlinespace[1pt]
\bottomrule
\end{tabular}
\end{adjustbox}
\vspace{-2pt}
\end{table*}
}

\section{Experimental Setup}
\label{sec:setup}

\paragraph{Datasets and tasks.}
We evaluated bioMoR on $8$ datasets: Genomap single-cell data
\cite{islam2023cartography}, Reactome-based cancer data \cite{gillespie2022reactome}, and TCGA
cohorts \cite{tcga2013pancancer} downloaded through Xena \cite{goldman2020xena}. Single-cell tasks
are multi-class cell-type classification; cancer tasks include binary classification and survival
prediction on HNSC (head and neck squamous cell carcinoma), UCEC (uterine corpus endometrial
carcinoma), COAD (colon adenocarcinoma), and KIRP (kidney renal papillary cell carcinoma)
\cite{tcga2013pancancer}, measured by Harrell's C-index
\cite{harrell1984regression}. Gene expression measures RNA abundance \cite{wang2009rnaseq},
mutation denotes a DNA-sequence alteration \cite{vogelstein2013cancer}, and copy-number variation
(CNV) denotes gains or losses of DNA segments \cite{feuk2006structural}. Single-cell inputs use gene
tokens. For cancer, we create each pathway token by pooling the features of genes assigned to that
Reactome pathway: mean pooling for gene expression, CNV and summed mutation burden. The pathway-based dataset Pancancer has two variants: PAN-2M uses mutation and
CNV, whereas PAN-3M additionally includes gene expression. Additional datasets with full results are given in the supplement.

\paragraph{Models and baselines.}
We first compare bioMoR with three relevant configurations of transformers: {\it Vanilla} is the
standard non-recursive Transformer. {\it Recursive} reuses one Transformer
block for a fixed number of steps. {\it MoR} adds adaptive recursion using 
expert-choice or token-choice routing \cite{bae2025mixture}. \textbf{bioMoR}
uses the same recursive backbone as MoR, but integrates biological knowledge
at the embedding, attention, and router sites. We also compare with representative and
state-of-the-art models, including Random Forest~\cite{breiman2001random},
Nearest-Centroid~\cite{tibshirani2002diagnosis}, scBiGNN~\cite{scbignn2023},
CNN(ei)~\cite{fu2020gene}, MOGONET~\cite{mogonet2021},
Pathformer~\cite{pathformer2023}, Geneformer~\cite{theodoris2023transfer},
scGPT~\cite{cui2024scgpt}, and scTransformer-style~\cite{sctransformer2024} baselines.
Please see the supplementary material for details about these models.

\paragraph{Implementation and evaluation.}
Hyperparameters are fixed by modality, with no per-dataset tuning. The single-cell configuration
follows the Genomap benchmark protocol \cite{islam2023cartography}: AdamW, learning rate
$10^{-3}$, weight decay $10^{-5}$, batch size $128$, $d_{\mathrm{model}}{=}96$, and $M{=}128$.
The pathway-based multi-omics configuration follows PATH \cite{howlader2026path}: AdamW, learning
rate $3\times10^{-4}$, weight decay $10^{-2}$, batch size $32$, $d_{\mathrm{model}}{=}128$, and
$M{=}256$. Both use $10\%$ linear warmup followed by cosine decay,
gradient clipping at $1.0$, dropout $0.1$, at most $100$ epochs, and patience $15$. Results use
paired $5$-fold stratified cross-validation with seed $42$; each fold holds out $20\%$ for testing
and $10\%$ of training for validation. The default is $K{=}4$, with $K\in\{2,3,4\}$ tested in RQ4.
Table~\ref{tab:selected_results} reports mean$\pm$SD. See the supplement for the full
hyperparameter configuration.

\section{Results}
\label{sec:main}

We organize the results around four research questions. RQ1 evaluates predictive gains, RQ2 tests
which biological-knowledge components matter, RQ3 measures whether the gains are computationally
efficient, and RQ4 studies routing, depth, and model configurations.


\noindent\textbf{RQ1: Does bioMoR improve predictive performance?}

In RQ1, we compare bioMoR with a controlled architecture ladder across eight datasets to determine
whether biology-guided routing improves predictive performance. Because these datasets are
class-imbalanced, Table~\ref{tab:selected_results} reports macro-F1 and balanced accuracy. The
architecture ladder consists of a vanilla Transformer, a fixed-depth recursive Transformer, and
biology-agnostic MoR. To align effective depth with the base Transformer, denoted Vanilla, all
recursive models in Table~\ref{tab:selected_results} use $K{=}4$. Both bioMoR routing variants
outperform these baselines on average for both metrics. The best bioMoR configuration improves
average macro-F1 by $8.2$ percentage points and balanced accuracy by $7.1$ percentage points over
the strongest non-bioMoR baseline. Additional datasets and their performances are provided in the supplementary
material.

\begin{figure}[H]
\centering
\includegraphics[width=0.92\columnwidth]{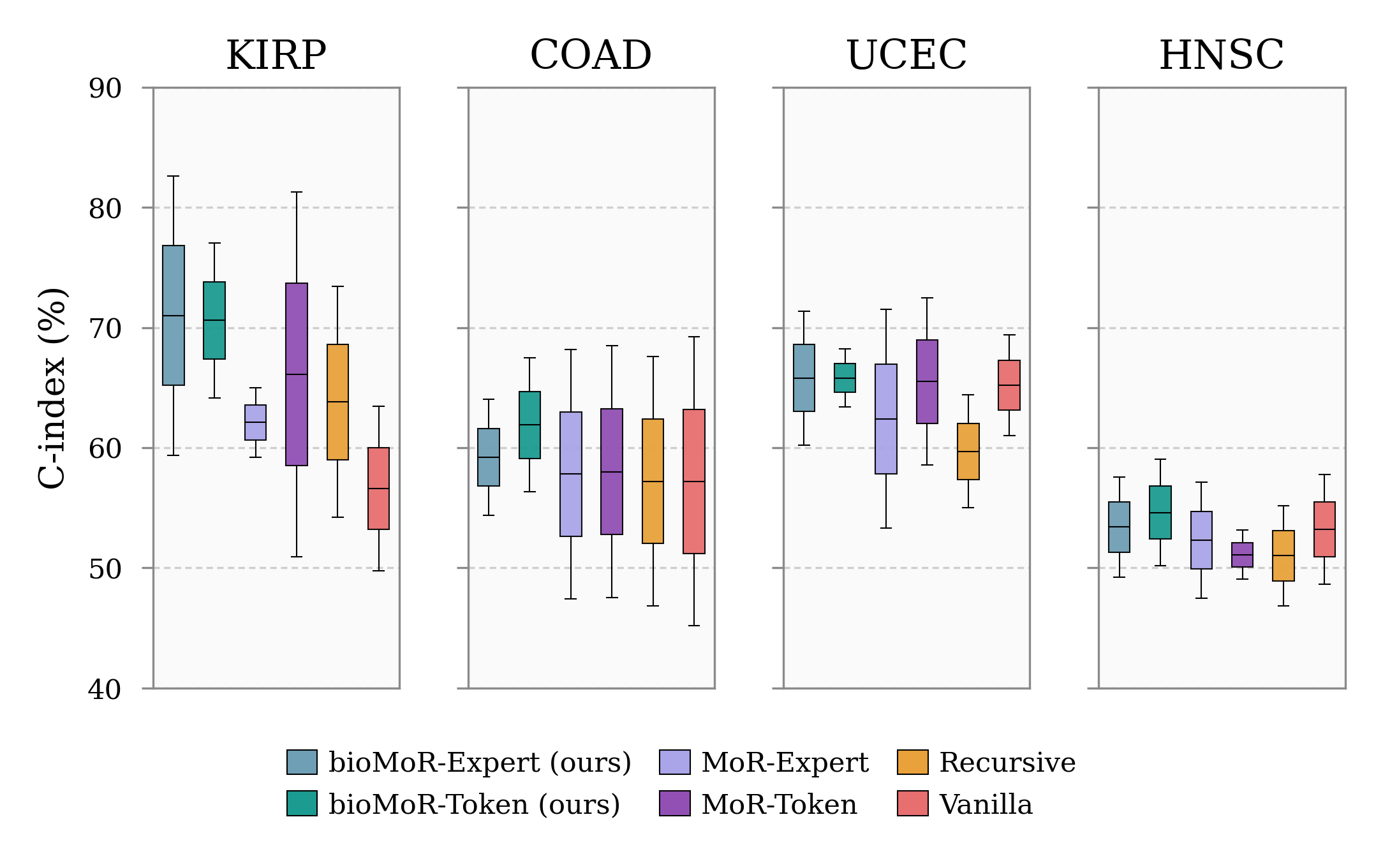}
\caption{\textbf{Survival prediction across TCGA cohorts.} Fold-level C-index distributions for
HNSC, UCEC, COAD, and KIRP. Higher values indicate better concordance between predicted risk and
observed survival outcomes.}
\label{fig:survival_cindex}
\end{figure}

\paragraph{Survival prediction.}
To test whether the architecture transfers beyond classification, we evaluate survival-risk
prediction on four TCGA cohorts. Figure~\ref{fig:survival_cindex} compares bioMoR-Expert and
bioMoR-Token with the Vanilla, Recursive, MoR-Expert, and MoR-Token baselines. The bioMoR variants are competitive
across HNSC, UCEC, COAD, and KIRP, showing that biology-guided adaptive recursion also transfers to
a time-to-event endpoint.

\begin{figure}[H]
\centering
\includegraphics[width=0.90\columnwidth]{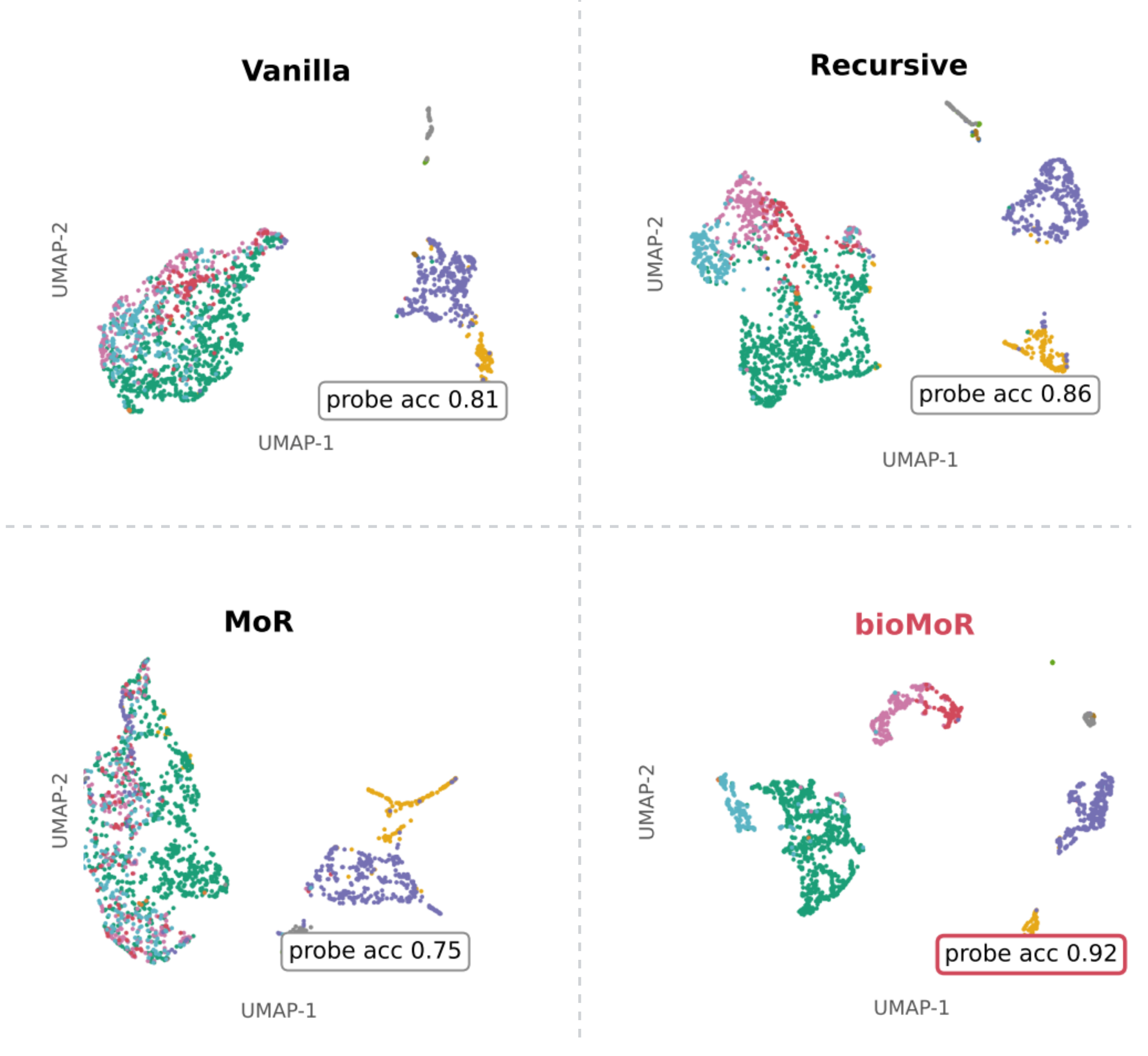}
\caption{\textbf{Frozen-embedding linear probe on Segerstolpe dataset.}
UMAP visualization of penultimate per-cell embeddings for Vanilla, Recursive, MoR, and bioMoR,
colored by cell type. }
\label{fig:baron_umap}
\end{figure}

Beyond predictive performance, we next examine whether these gains are reflected in the learned
representations. Figure~\ref{fig:baron_umap} visualizes the frozen embeddings on Segerstolpe using
UMAP \cite{mcinnes2018umap} and evaluates their linear separability with a linear probe
\cite{alain2016linear}.
As shown in Figure~\ref{fig:baron_umap}, bioMoR forms cleaner cell-type groups in the frozen
embedding space, and the linear probe performs best on these embeddings. This suggests that bioMoR
improves the learned representation rather than only the final prediction layer. Additional
visualizations are provided in the supplement.

\begin{figure}[H]
\centering
\includegraphics[width=0.90\columnwidth]{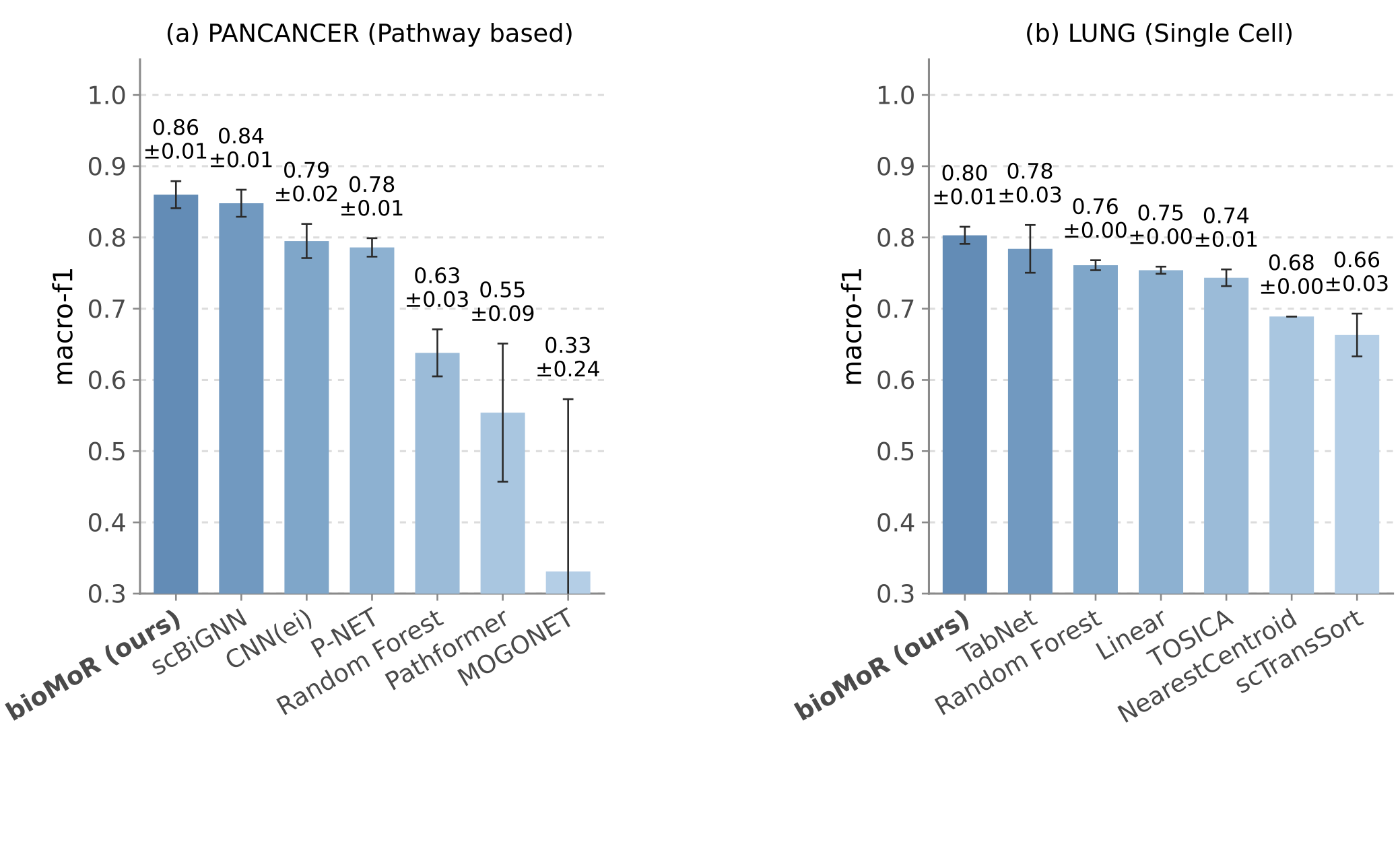}
\caption{\textbf{Comparison with external baselines.} Mean macro-F1$\pm$SD for bioMoR and
dataset-appropriate baselines on (a) pathway-based PAN-2M and (b) single-cell Lung under the
matched five-fold evaluation setting.}
\label{fig:sota_compare}
\end{figure}

Finally, to assess whether the gains extend beyond our internal architecture ladder,
Figure~\ref{fig:sota_compare} compares bioMoR with dataset-appropriate external state-of-the-art
baselines on PAN-2M and Lung as representative datasets. The full set of baselines is described
under \textit{Models and baselines}.
bioMoR obtains the highest mean macro-F1 in both panels: $0.86{\pm}0.01$ on PAN-2M, compared
with $0.84{\pm}0.01$ for scBiGNN, and $0.80{\pm}0.01$ on Lung, compared with
$0.78{\pm}0.03$ for TabNet. Additional results for external baselines are provided in the
supplement.

\FloatBarrier

\subsection{Ablation Results}
\label{sec:interaction}
\begin{figure}[t]
\centering
\IfFileExists{figs/ablation_bar.pdf}{\includegraphics[width=0.90\columnwidth]{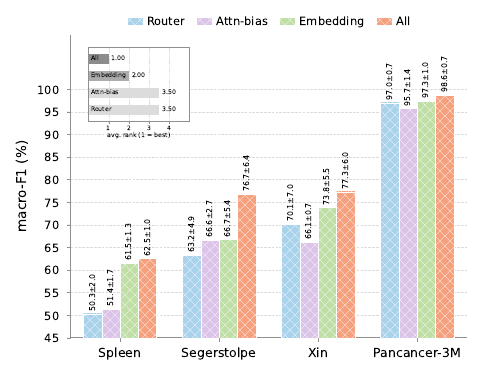}}{\emph{(ablation bar chart renders here)}}
\caption{\textbf{Injection-site ablation.} Macro-F1 across Spleen, Segerstolpe, Xin, and
Pancancer-3M with biological-knowledge injection at the embedding, attention-bias, router, or all three sites.
All experiments use expert-choice routing with $K{=}4$. The inset shows the average rank across
datasets ($1$ is best).}
\label{fig:ablation_bar}
\end{figure}

\paragraph{RQ2: In which component(s), the biological knowledge is most useful?}
To evaluate the contribution of each biological injection site, we conduct an ablation study on the
token embedding, attention bias, and router. We ran all the 8 datasets, and due to the space, we showed results for four representative datasets in Figure~\ref{fig:ablation_bar}. Our results indicate that integrating
biological knowledge at all three sites achieves the highest macro-F1. Among the single-site variants, embedding injection performs best, followed by router-only and attention-bias-only injection. Overall, these results show that
the three injection sites provide complementary information, with the strongest performance obtained
when all three are used together.

\subsection{Efficiency}
\label{sec:efficiency}
\begin{figure}[t]
\centering
\IfFileExists{figs/t_cell_efficiency_600.png}{\includegraphics[width=0.90\columnwidth]{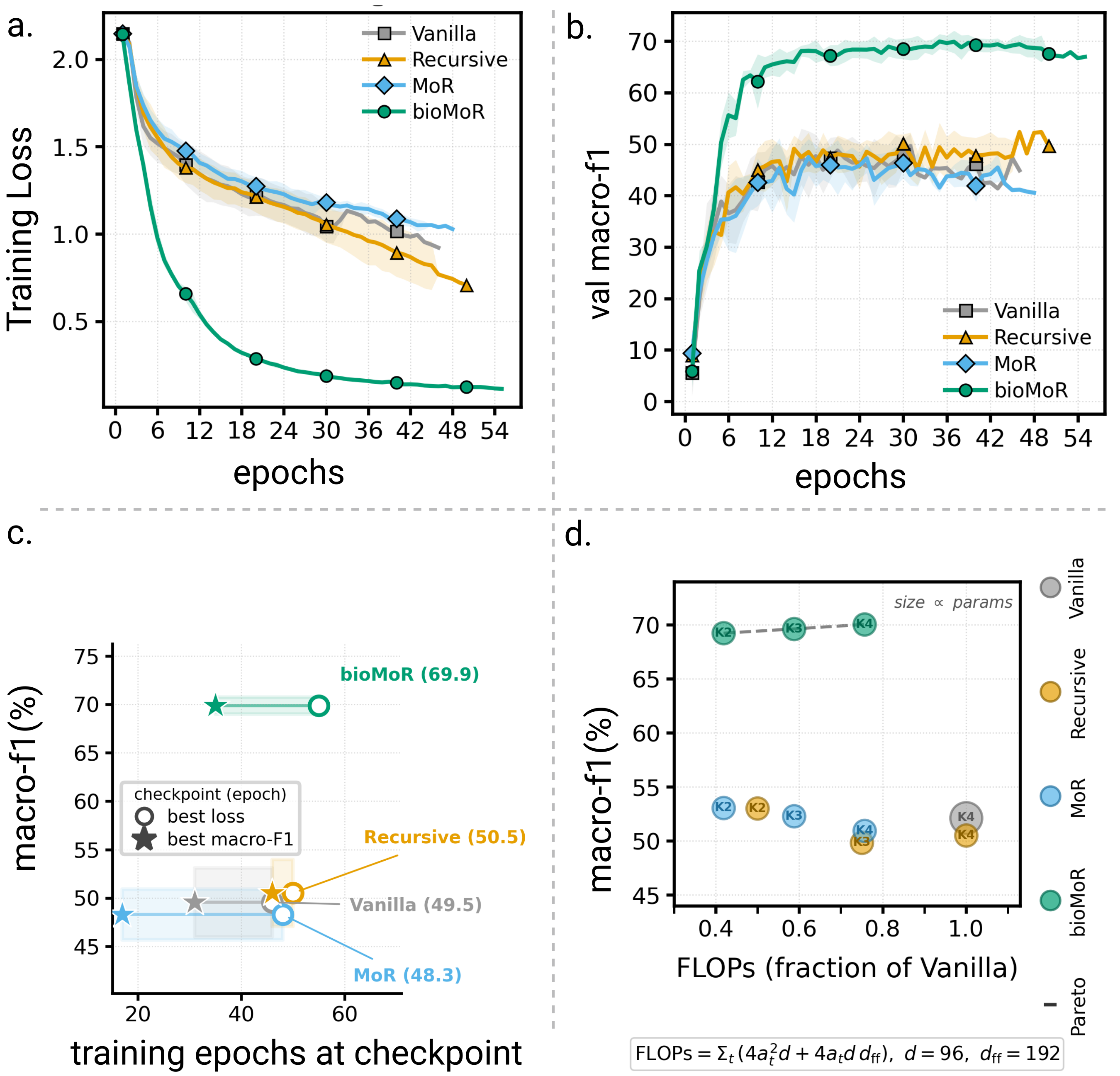}}{\emph{(T-cell efficiency figure renders here)}}
\caption{\textbf{Training and computational efficiency on T-cell.} (a) Training loss and
(b) validation macro-F1 over epochs. (c) Test macro-F1 and training epoch at checkpoints selected
by lowest validation loss ($\circ$) or highest validation macro-F1 ($\star$). (d) Test macro-F1
versus FLOPs for $K{=}2,3,4$; bubble size represents parameter count.}
\label{fig:tcell_efficiency}
\end{figure}



\paragraph{RQ3: What are the training and FLOP gains?}
RQ3 asks whether bioMoR improves training behavior while reducing computational cost. We evaluate computational efficiency across all datasets and provide the complete results in the supplementary material. Due to the main-paper space limit, we present T-cell as a representative dataset from Table~\ref{tab:selected_results}. With expert-choice routing at $K{=}4$, we compare learning curves and checkpoint-selected test performance across the internal architecture ladder (Figure~\ref{fig:tcell_efficiency}a--c). In panel (a), bioMoR reduces the training loss faster and reaches the lowest final value. In panel (b), it achieves nearly $70\%$ validation macro-F1, whereas the baselines remain near $40$--$50\%$. Panel (c) confirms this advantage under both checkpoint-selection rules, with bioMoR achieving the highest test macro-F1 of $69.9\%$.

Panel (d) compares predictive performance and computational cost. Each circle represents one model at a specific recursion depth. Models closer to the upper-left are preferable because they achieve higher macro-F1 with fewer FLOPs. bioMoR consistently maintains approximately $69$--$70\%$ macro-F1 across all recursion depths while requiring fewer FLOPs than Vanilla. At $K{=}2$, bioMoR uses $58\%$ fewer FLOPs than Vanilla.

\subsection{Configuration Studies}
\label{sec:qualitative}

\paragraph{RQ4: Which configuration choices matter?}
RQ4 examines whether bioMoR depends on a specific configuration. We evaluate all datasets and report the complete results in the supplement; due to space limits, Table~\ref{tab:rq4_config} shows T-cell as a representative case. bioMoR remains strong under both routing policies and across $K{=}2,3,4$, while Vanilla, Recursive, and MoR perform substantially worse. Thus, biological guidance drives the main gain, while routing policy and depth adjust the accuracy--efficiency trade-off.


\begin{table}[H]
\centering
\scriptsize
\setlength{\tabcolsep}{2pt}
\renewcommand{\arraystretch}{1.00}
\caption{\textbf{RQ4 T-cell configuration slice.} T-cell macro-F1 from Supplementary
Table~1. Each cell reports normalized FLOPs / macro-F1 (mean$\pm$SD).}
\label{tab:rq4_config}
\begin{tabular}{@{}llccc@{}}
\toprule
Model & Routing & \multicolumn{3}{c}{Normalized FLOPs / macro-F1} \\
\cmidrule(lr){3-5}
& & $K{=}2$ & $K{=}3$ & $K{=}4$ \\
\midrule
Vanilla & -- & -- & -- & $1.00/52.1{\scriptstyle\pm3.5}$ \\
Recursive & -- & $0.50/53.0{\scriptstyle\pm2.0}$ &
$0.75/49.8{\scriptstyle\pm4.4}$ & $1.00/50.5{\scriptstyle\pm3.5}$ \\
\midrule
MoR & Expert & $0.42/50.6{\scriptstyle\pm4.9}$ &
$0.59/52.3{\scriptstyle\pm2.4}$ & $0.76/50.9{\scriptstyle\pm4.7}$ \\
MoR & Token & $0.42/50.7{\scriptstyle\pm2.2}$ &
$0.52/55.8{\scriptstyle\pm3.0}$ & $0.56/50.4{\scriptstyle\pm2.2}$ \\
\midrule
\textbf{bioMoR} & Expert & $0.42/\mathbf{69.2{\scriptstyle\pm0.7}}$ &
$0.59/\mathbf{69.6{\scriptstyle\pm0.6}}$ &
$0.76/\mathbf{70.0{\scriptstyle\pm1.9}}$ \\
\textbf{bioMoR} & Token & $0.42/\mathbf{68.6{\scriptstyle\pm1.7}}$ &
$0.52/\mathbf{68.6{\scriptstyle\pm0.9}}$ &
$0.56/\mathbf{69.8{\scriptstyle\pm1.2}}$ \\
\bottomrule
\end{tabular}
\end{table}


Tables~\ref{tab:rq4_marker_budget} and~\ref{tab:rq4_width} examine the marker budget $M$ and model
width $d$, respectively. Mean macro-F1 remains within $71.8$--$73.5$ as $M$ increases from $128$ to
$2048$: T-cell and Spleen improve, Lung remains stable, and Muraro and Segerstolpe show no consistent
trend. As $d$ increases from $96$ to $352$, mean macro-F1 rises from $70.5$ to $72.7$, although the
change is not monotonic; Lung, T-cell, and Spleen remain stable, Muraro generally improves, and
Segerstolpe, BLCA, and PAN-2M vary across settings. Overall, bioMoR's improvement does not depend on a specific configuration;
the performance gain is primarily driven by biological knowledge.

\begin{table}[H]
\centering
\scriptsize
\setlength{\tabcolsep}{3pt}
\renewcommand{\arraystretch}{1.00}
\caption{\textbf{RQ4 marker-budget headroom.} bioMoR macro-F1 (mean$\pm$SD) across the eight
datasets shown in Table~\ref{tab:selected_results}.}
\label{tab:rq4_marker_budget}
\vspace{2pt}
\begin{tabular}{lccccc}
\toprule
Dataset & $M{=}128$ & $M{=}256$ & $M{=}512$ & $M{=}1024$ & $M{=}2048$ \\
\midrule
Lung & $79.1{\scriptstyle\pm1.2}$ & $80.2{\scriptstyle\pm0.8}$ & $79.8{\scriptstyle\pm1.9}$ & $81.3{\scriptstyle\pm2.0}$ & $81.1{\scriptstyle\pm1.9}$ \\
Muraro & $75.3{\scriptstyle\pm6.4}$ & $72.4{\scriptstyle\pm2.7}$ & $68.3{\scriptstyle\pm3.3}$ & $75.7{\scriptstyle\pm4.5}$ & $73.8{\scriptstyle\pm6.0}$ \\
Segerstolpe & $71.4{\scriptstyle\pm5.3}$ & $70.3{\scriptstyle\pm4.1}$ & $73.8{\scriptstyle\pm5.4}$ & $69.4{\scriptstyle\pm14.7}$ & $68.3{\scriptstyle\pm3.3}$ \\
Spleen & $58.2{\scriptstyle\pm0.9}$ & $58.6{\scriptstyle\pm1.6}$ & $59.9{\scriptstyle\pm0.8}$ & $61.1{\scriptstyle\pm2.3}$ & $61.6{\scriptstyle\pm4.0}$ \\
T-cell & $68.9{\scriptstyle\pm1.0}$ & $69.9{\scriptstyle\pm0.3}$ & $71.5{\scriptstyle\pm0.4}$ & $71.8{\scriptstyle\pm2.3}$ & $73.0{\scriptstyle\pm0.3}$ \\
\midrule
BLCA & $44.9{\scriptstyle\pm7.0}$ & $41.5{\scriptstyle\pm6.7}$ & $42.6{\scriptstyle\pm10.0}$ & $45.6{\scriptstyle\pm4.9}$ & $44.6{\scriptstyle\pm6.4}$ \\
PAN-2M & $85.5{\scriptstyle\pm1.9}$ & $85.0{\scriptstyle\pm0.9}$ & $85.7{\scriptstyle\pm2.1}$ & $86.3{\scriptstyle\pm1.3}$ & $84.4{\scriptstyle\pm0.9}$ \\
PAN-3M & $96.7{\scriptstyle\pm1.4}$ & $96.1{\scriptstyle\pm0.7}$ & $96.3{\scriptstyle\pm0.5}$ & $96.4{\scriptstyle\pm1.3}$ & $95.5{\scriptstyle\pm0.8}$ \\
\midrule
\textbf{Mean} & $\mathbf{72.5}$ & $\mathbf{71.8}$ & $\mathbf{72.2}$ & $\mathbf{73.5}$ & $\mathbf{72.8}$ \\
\bottomrule
\end{tabular}
\end{table}

\begin{table}[H]
\centering
\scriptsize
\setlength{\tabcolsep}{2.5pt}
\renewcommand{\arraystretch}{1.00}
\caption{\textbf{RQ4 model-width sweep (5-fold CV).} bioMoR macro-F1 (mean$\pm$SD over the unified
five folds) across $d_{\mathrm{model}}\in\{96,136,192,272,352\}$ at fixed marker budget $M{=}256$,
on five single-cell and three multi-omics datasets.}
\label{tab:rq4_width}
\vspace{2pt}
\begin{tabular}{lccccc}
\toprule
Dataset & $d{=}96$ & $d{=}136$ & $d{=}192$ & $d{=}272$ & $d{=}352$ \\
\midrule
Lung & $80.2{\scriptstyle\pm1.4}$ & $79.6{\scriptstyle\pm0.7}$ & $80.6{\scriptstyle\pm1.8}$ & $81.6{\scriptstyle\pm0.9}$ & $80.4{\scriptstyle\pm1.0}$ \\
Muraro & $79.7{\scriptstyle\pm2.3}$ & $85.4{\scriptstyle\pm5.9}$ & $82.6{\scriptstyle\pm6.6}$ & $87.4{\scriptstyle\pm5.6}$ & $88.6{\scriptstyle\pm5.6}$ \\
Segerstolpe & $76.0{\scriptstyle\pm3.4}$ & $75.2{\scriptstyle\pm6.7}$ & $79.9{\scriptstyle\pm4.2}$ & $74.8{\scriptstyle\pm6.6}$ & $80.5{\scriptstyle\pm6.9}$ \\
Spleen & $60.6{\scriptstyle\pm0.6}$ & $60.3{\scriptstyle\pm1.6}$ & $61.7{\scriptstyle\pm1.3}$ & $60.3{\scriptstyle\pm0.4}$ & $60.9{\scriptstyle\pm0.9}$ \\
T-cell & $69.5{\scriptstyle\pm1.0}$ & $69.2{\scriptstyle\pm2.0}$ & $70.1{\scriptstyle\pm1.3}$ & $69.7{\scriptstyle\pm0.9}$ & $69.2{\scriptstyle\pm1.9}$ \\
\midrule
BLCA & $42.2{\scriptstyle\pm2.1}$ & $43.1{\scriptstyle\pm8.4}$ & $42.8{\scriptstyle\pm9.3}$ & $44.6{\scriptstyle\pm5.3}$ & $42.3{\scriptstyle\pm3.6}$ \\
PAN-2M & $85.7{\scriptstyle\pm0.5}$ & $86.4{\scriptstyle\pm1.9}$ & $86.9{\scriptstyle\pm1.6}$ & $84.6{\scriptstyle\pm2.5}$ & $87.1{\scriptstyle\pm1.6}$ \\
PAN-3M & $96.7{\scriptstyle\pm1.4}$ & $96.1{\scriptstyle\pm0.7}$ & $96.3{\scriptstyle\pm0.5}$ & $96.4{\scriptstyle\pm1.3}$ & $95.5{\scriptstyle\pm0.8}$ \\
\midrule
\textbf{Mean} & $\mathbf{73.8}$ & $\mathbf{74.4}$ & $\mathbf{75.1}$ & $\mathbf{74.9}$ & $\mathbf{75.6}$ \\
\bottomrule
\end{tabular}
\end{table}

\begin{figure}[t]
\centering
\IfFileExists{figs/case_study.pdf}{\includegraphics[width=0.90\columnwidth]{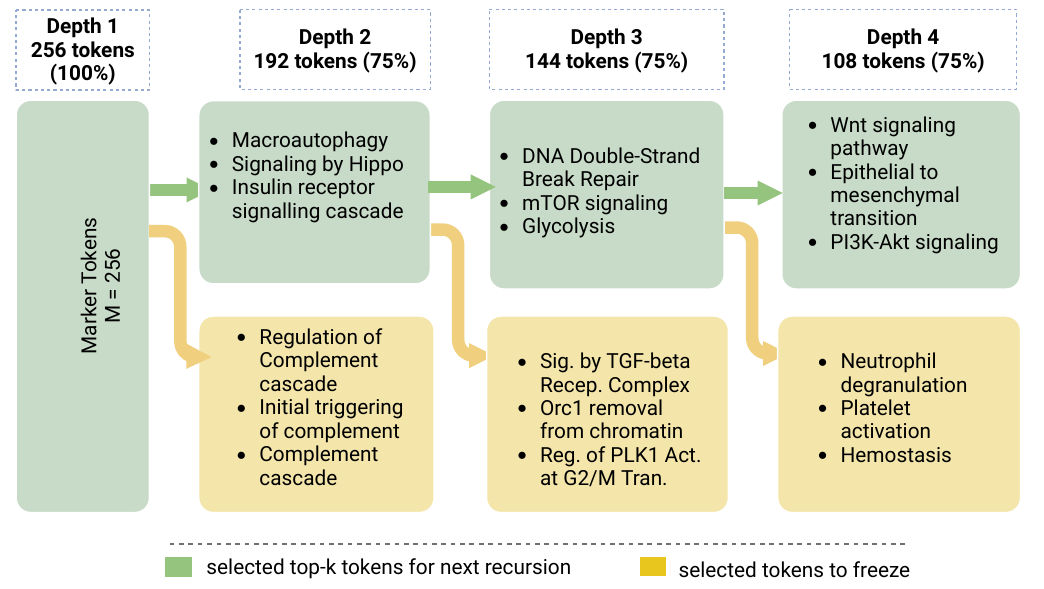}}{\emph{(pathway-prioritization case-study figure renders here)}}
\caption{\textbf{Pathway-prioritization case study.} Green boxes give examples of pathway tokens
assigned to receive deeper computation; yellow boxes give examples that stop at the corresponding
depth and retain their current representations.}
\label{fig:case_study}
\end{figure}

\subsection{Biological Case Study: Pathways Prioritized for Deeper Computation}
\label{sec:case_study}
We use PAN-2M to examine which pathway tokens bioMoR processes more deeply for localized-versus-metastatic cancer classification. Because each token represents a named biological pathway, the routing decisions can be compared with existing cancer studies.

At each recursion, the router sends the top $75\%$ of active pathways to the next depth, while the remaining $25\%$ stop. Thus, pathways reaching deeper levels receive more computation. In Figure~\ref{fig:case_study}, green boxes indicate continuing pathways and yellow boxes indicate stopped pathways.

Pathways receiving the deepest computation include Wnt signaling, epithelial-to-mesenchymal transition (EMT), and PI3K--Akt signaling, all of which are linked to cancer invasion or metastasis \cite{nguyen2009wnt,grasset2022emt,thibault2021pi3k}. This agreement with prior studies indicates that bioMoR prioritizes pathways known to be relevant to metastatic cancer.




\FloatBarrier

\section{Conclusion}
We developed bioMoR, a biological-knowledge-guided Mixture-of-Recursions framework and, to the best
of our knowledge, the first application of MoR to genomic learning. We integrate biological knowledge
at three sites: to refine token embeddings, bias self-attention, and route each token to an appropriate
recursion depth. This design directly addresses a central computational mismatch in omics modeling:
most features should not require the same amount of deep computation, and the decision about where to
spend computation should be informed by molecular structure rather than by token embeddings alone.
Our evaluation demonstrates that bioMoR significantly improves predictive performance while being
more computationally efficient, as shown by fewer model parameters and reduced FLOPs. These
improvements are consistent across diverse genomic datasets and the extensive configurations we
evaluated. Overall, our results suggest that biological structure is useful not only as knowledge for
representation learning, but also as a control signal for adaptive computation in genomic models.

\section*{Acknowledgments}
This work is partially supported by the National Science Foundation under Grant No.~2152117. The
research reported in this paper is partially supported by the HPC@ISU equipment at Iowa State
University, some of which has been purchased through funding provided by NSF under MRI grants
number 1726447 and MRI2018594.

\FloatBarrier

\begingroup
\small
\bibliographystyle{aaai}
\bibliography{refs}
\endgroup

\clearpage


\section*{Supplementary material}

# Supplementary Material
# bioMoR: Biology-Guided Mixture-of-Recursions for Effective Genomic Learning

**Koushik Howlader**[1], **Tirtho Roy**[1], **Md Tauhidul Islam**[2], **Wei Le**[1]
[1]Iowa State University, Iowa, USA
[2]Stanford University, CA, USA

## Appendix A: Extended Related Work

**Genomic representation learning.** Genomic samples contain thousands of correlated and often sparse molecular measurements. Single-cell Transformers such as scBERT (Yang et al. 2022), Geneformer (Theodoris et al. 2023), and scGPT (Cui et al. 2024) show that large-scale pretraining can support cell annotation, perturbation analysis, and transfer to network-biology tasks; scFoundation further studies scaling to larger cell collections (Hao et al. 2024). These models establish the value of learned gene representations, but their standard backbones normally apply the same sequence of layers to every retained gene. Related feature-selection methods focus computation on highly variable or marker genes (Luecken and Theis 2019; Ianevski, Giri, and Aittokallio 2022), while Genomap exploits structured gene relationships to construct compact single-cell representations (Islam and Xing 2023). Selection, however, does not determine how deeply each retained feature should be processed. bioMoR separates these decisions: it first forms interpretable marker or pathway tokens and then assigns token-specific recursive depth.

**Biological priors in omics models.** Biological structure enters omics models through curated hierarchies, pathways, and interaction graphs. P-NET constrains a cancer model by gene/pathway organization (Elmarakeby et al. 2021); TOSICA uses pathway-informed attention for interpretable cell annotation (Chen et al. 2023); Pathformer models pathway interactions in multi-omics data (Liu et al. 2024); and MOGONET and scBiGNN use graph structure for multi-omics and single-cell prediction (Wang et al. 2021; Yang et al. 2023). These studies show that molecular relations improve representation quality and interpretability. Most use biology to define connectivity, aggregate features, or bias attention, rather than to control whether a token is processed again. bioMoR uses one knowledge source at three sites: graph smoothing refines embeddings, structural bias guides attention, and a neighborhood-aware router determines recursion depth.



**Parameter-efficient depth and adaptive computation.** Universal Transformers repeatedly apply one transition block (Dehghani et al. 2019), while ALBERT shows that cross-layer sharing can reduce parameters (Lan et al. 2020); both nevertheless use fixed computation. Adaptive Computation Time learns when a recurrent state should halt (Graves 2016), sparsely gated mixture-of-experts routes tokens among experts (Shazeer et al. 2017), and Mixture-of-Depths allocates different amounts of layer computation under a budget (Raposo et al. 2024). Mixture-of-Recursions combines adaptive routing with a shared recursive block, decoupling maximum depth from parameter count (Bae et al. 2025). These routers are general-purpose and must infer biological relationships only from task data.

**Positioning of bioMoR.** bioMoR connects biology-informed genomic learning with adaptive recursion. It is not a replacement for foundation-model pretraining or a new pathway database; it adds a biology-guided computation policy to a compact Transformer. Compared with fixed biological architectures, it allocates unequal depth across tokens. Compared with biology-agnostic MoR, it uses gene co-expression or Reactome relations to inform representations, attention, and routing while preserving a no-op path for the biological correction. Thus, bioMoR addresses the main paper's central mismatch: genomic tokens differ in both predictive relevance and the relational computation needed to resolve their contribution.

# Appendix B: Dataset and Baseline Details

## Dataset Selection

The main paper reports a representative eight-dataset classification subset. The complete classification suite contains 14 datasets: 8 Genomap single-cell datasets (Islam and Xing 2023), 3 Reactome multi-omics cohorts (Gillespie et al. 2022), and 3 pan-cancer TCGA settings (The Cancer Genome Atlas Research Network et al. 2013). Table 1 summarises the dataset name, sample count, feature type, and classification type.

| Dataset | Samples | Feature type | Classification type |
|---|---|---|---|
| *Single-cell genomap datasets* | | | |
| Baron | 8,569 | scRNA-seq Expression | Multi-class (13 classes) |
| Lung | 57,020 | scRNA-seq Expression | Multi-class (25 classes) |
| Muraro | 2,122 | scRNA-seq Expression | Multi-class (9 classes) |
| Oesophagus | 87,947 | scRNA-seq Expression | Multi-class (18 classes) |
| Segerstolpe | 2,127 | scRNA-seq Expression | Multi-class (11 classes) |
| Spleen | 94,129 | scRNA-seq Expression | Multi-class (28 classes) |
| T-cell | 24,007 | scRNA-seq Expression | Multi-class (7 classes) |
| Xin | 1,492 | scRNA-seq Expression | Multi-class (4 classes) |
| *Cancer pathway-token datasets* | | | |
| Prostate | 1,011 | Mutation+CNV+Expression | Binary classification |
| BLCA (bladder) | 404 | Mutation+CNV+Expression | Binary classification |
| STAD (stomach) | 414 | Mutation+CNV+Expression | Binary classification |
| Pan-cancer mutation+CNV | 8,893 | CNV+Mutation | Binary classification |
| Pan-cancer expression | 8,586 | Expression | Binary classification |
| Pan-cancer tri-modal | 8,586 | CNV+Mutation+Expression | Binary classification |

Table 1: Dataset summary. Cancer datasets use Reactome pathway-token inputs; the pan-cancer cohorts evaluate three feature views: CNV+mutation, expression only, and CNV+mutation+expression.

## Baseline Selection

| Model or baseline | Role in comparison | Reference |
|---|---|---|
| Vanilla Transformer | Non-recursive internal baseline | (Vaswani et al. 2017) |
| Recursive Transformer | Weight-shared fixed-depth baseline | (Dehghani et al. 2019; Lan et al. 2020) |
| MoR | Biology-agnostic adaptive-recursion baseline | (Bae et al. 2025) |
| Random Forest | Classical tree-based classifier | (Breiman 2001) |
| Nearest-Centroid | Classical centroid-based classifier | (Tibshirani et al. 2002) |
| TabNet | Attentive tabular-learning baseline | (Arik and Pfister 2021) |
| TOSICA | Pathway-aware single-cell Transformer | (Chen et al. 2023) |
| scBiGNN | Single-cell graph neural baseline | (Yang et al. 2023) |
| MOGONET | Multi-omics graph neural baseline | (Wang et al. 2021) |
| Pathformer | Pathway-informed transformer baseline | (Liu et al. 2024) |
| Geneformer | Gene-vocabulary foundation model baseline | (Theodoris et al. 2023) |
| scGPT | Single-cell foundation model baseline | (Cui et al. 2024) |
| scTransformer-style | Transformer with gene-regulatory priors | (Milia et al. 2026) |
| CNN(ei) | Expression-image convolutional baseline | (Fu et al. 2020) |

Table 2: Model and baseline references used in the experimental comparisons. Internal baselines test architectural changes under the same bioMoR input pipeline; external baselines are included when their input assumptions match the corresponding dataset type.

# Appendix C: Experimental Setup

**Hyperparameter configuration.** Fixed training configuration (no per-dataset tuning); each modality uses one configuration held fixed across all its datasets and variants. *LR schedule*: linear warmup over the first 10% of epochs (1% → 100% of the base rate), then cosine annealing. *Early stopping* on validation macro-F1. *Protocol*: 5-fold stratified cross-validation (seed 42; 20% test, 10%-of-train validation), per-gene $z$-scoring fit on the train fold only; class-weighted cross-entropy loss. $M$ counts learnable marker tokens (single-cell) or fixed Reactome pathway tokens (multi-omics).

| Setting | Single-cell | Multi-omics |
|---|---|---|
| Optimiser | AdamW | AdamW |
| Base learning rate | $10^{-3}$ | $3 \times 10^{-4}$ |
| Weight decay | $10^{-5}$ | $10^{-2}$ |
| LR schedule | warmup → cosine | warmup → cosine |
| Gradient clip ($\ell_2$) | 1.0 | 1.0 |
| Dropout | 0.1 | 0.1 |
| Batch size | 128 | 32 |
| Max epochs | 100 | 100 |
| Early-stop patience | 15 | 15 |
| Width $d_{\text{model}}$ | 96 | 128 |
| FFN width $d_{\text{ff}}$ | 192 | 256 |
| Token budget $M$ | 128 | 256 |
| Recursion depth $K$ | 4 | 4 |
| CV folds / seed | 5 / 42 | 5 / 42 |

# Appendix D: Extended Method

## bioMoR

**Overview and notation.** For one sample, $x$ denotes the measured genomic input: a gene-expression vector for single-cell data or aligned mutation, copy-number, and expression channels for multi-omics data. Tokenization converts $x$ into $T = [t_1, \ldots, t_M]^\top \in \mathbb{R}^{M \times d}$, where $M$ is the number of marker-gene or pathway tokens, $d$ is the hidden width, and $t_m$ is the initial representation of token $m$. The symbol $K$ denotes the largest permitted recursion depth. At recursion step $t$, $H^{(t)} \in \mathbb{R}^{M \times d}$ stores the current token states, while $\mathcal{A}_t \subseteq \{1, \ldots, M\}$ contains only the tokens that remain active. A token removed from $\mathcal{A}_t$ keeps its latest state, so routing changes computation without deleting its information from the final pooled representation.

**Algorithm 1: bioMoR forward pass.**

**Input:** $x$, biological knowledge $B$
  token budget $M$, max depth $K$
$T \leftarrow$ marker-gene or pathway tokens from $x$
$A \leftarrow$ row-normalized fixed knowledge matrix from $B$
$H \leftarrow (1-\lambda)T + \lambda AT; \quad \mathcal{A} \leftarrow \{1, \ldots, M\}$
**for** $t = 1, \ldots, K$ **while** $\mathcal{A} \neq \emptyset$ **do**
  $H_{\mathcal{A}} \leftarrow f_\theta(H_{\mathcal{A}}; A)$ *reuse the same block*
  $r_{\mathcal{A}} \leftarrow$ biology-aware router scores
  $\mathcal{A}_{\text{next}} \leftarrow$ tokens routed to the next step
  tokens in $\mathcal{A} \setminus \mathcal{A}_{\text{next}}$ stop and keep current states
  $\mathcal{A} \leftarrow \mathcal{A}_{\text{next}}$ *return kept tokens to recursion*
**end for**
$\hat{y} \leftarrow \text{classifier}\big(\text{pool}(H)\big)$
**Train:** minimize class-weighted CE plus auxiliary losses
**Output:** $\hat{y}$, markers/pathways, recursion depths $d_m$

**Reading Algorithm 1.** $B$ is fixed biological knowledge and $A$ is its normalized form. The shared Transformer $f_\theta$ has trainable parameters $\theta$ and is reused at every step, so increasing $K$ increases possible computation but not the number of block parameters. The score vector $r_{\mathcal{A}}$ ranks or gates the active tokens; $\mathcal{A}_{\text{next}}$ is the subset sent through another application of $f_\theta$. After routing stops, $\text{pool}(H)$ combines all final token states, the classifier produces the predicted label distribution $\hat{y}$, and $d_m \in \{1, \ldots, K\}$ records how many recursive updates token $m$ received. Consequently, $d_m$ is both a compute-allocation variable and an interpretable summary of which marker or pathway required more refinement.

**Biological-knowledge construction.** Following Genomap (Islam and Xing 2023), we construct a gene *co-expression* matrix that summarizes pairwise gene relationships. Its entries are partial correlations computed from the inverse covariance of the gene-expression matrix and restricted to the selected markers; this matrix forms the single-cell biological knowledge $B$. For multi-omics data, we use curated Reactome pathway relationships and treat them as undirected: $B_{ij} = B_{ji} = 1$ when pathways $i$ and $j$ are related, and both entries are zero otherwise (Gillespie et al. 2022).

Let $D$ be the diagonal matrix of row sums of $B$. We use the fixed, row-normalized matrix

$$A = D^{-1}B. \tag{1}$$

$B_{ij}$ measures or indicates the biological relation between tokens $i$ and $j$; $D_{ii} = \sum_j B_{ij}$ is token $i$'s total connection weight; and $A_{ij} = B_{ij}/D_{ii}$ is the fraction of its neighborhood assigned to token $j$. Thus, row $i$ of $AH$ is a weighted average of the current states of token $i$'s biological neighbors. Row normalization prevents highly connected genes or pathways from producing a larger message merely because they have more neighbors. Isolated tokens, if present, receive no neighbor message and therefore rely on their own representation through the residual mixture in Site 1.

The same biological knowledge is used for embedding smoothing (Site 1), attention bias (Site 2), and depth routing (Site 3).

**Site 1: biology-guided embedding smoothing.** Before recursion, we smooth the tokens using the biological-knowledge graph $A$, so that information of biologically related tokens is also integrated into the embedding of the token:

$$\widetilde{T} = (1-\lambda)T + \lambda AT, \tag{2}$$

where $\lambda$ is learned. This denoises each marker/pathway token toward its biological neighbours before the repeated recursive updates.

Here, $\widetilde{T}$ is the representation passed to the first recursive step, $AT$ is the neighborhood-averaged representation, and $\lambda$ controls the mixture. When $\lambda = 0$, Site 1 reduces to the original token representation; larger positive values give biological neighbors more influence. This site can improve robustness when related genes or pathways carry complementary signal, but an excessively strong graph contribution can blur genuinely distinct tokens. Learning $\lambda$ lets validation loss determine the useful balance rather than fixing it manually.

**Site 2: attention bias.** Inside the recursive block, the *same* biological-knowledge graph $A$ biases self-attention:

$$\text{Attn}(Q, K, V) = \text{softmax}\left(\frac{QK^\top}{\sqrt{d}} + \lambda_{\text{attn}}\mathbf{1}[A > 0]\right)V. \tag{3}$$

This encourages biologically related tokens to exchange information during refinement.

$Q$, $K$, and $V$ are the query, key, and value matrices computed from the current token states; the $K$ in this equation means the *key matrix*, not maximum recursion depth. The factor $\sqrt{d}$ keeps dot-product magnitudes stable as the hidden width changes. The binary mask $\mathbf{1}[A > 0]$ equals one for a biological edge and zero otherwise, while $\lambda_{\text{attn}}$ is the learned strength of the added logit bias. Therefore, the graph does not force attention: it raises the prior logit of related pairs, and the data-dependent term $QK^\top$ can still override it. This can concentrate information exchange on plausible relationships without preventing the model from learning task-specific nonlocal interactions.

**Site 3: biology-aware routing.** The original MoR router uses the current token representation $h_m^{(t)}$ to decide whether token $m$ should continue to the next recursion step. bioMoR

keeps this original score and adds a learned correction based on the token's biological neighborhood:

$$\tilde{r}_m^{(t)} = \frac{w_r^\top h_m^{(t)}}{\tau_r} + \sigma(\gamma_t)\phi_\psi\Big([\, AH^{(t)},\ H^{(t)} - AH^{(t)}\,]_m\Big). \quad (4)$$

The first term is the original MoR routing score. In the second term, $AH^{(t)}$ summarizes information from biologically related tokens, while $H^{(t)} - AH^{(t)}$ shows how the token differs from its biological neighborhood. The small network $\phi_\psi$ uses these two signals to adjust the routing score, and $\sigma(\gamma_t)$ controls how much correction is added at recursion step $t$. Because $\phi_\psi$ is initialized to output zero, the correction is zero at the start of training and bioMoR initially behaves like biology-agnostic MoR. During training, the correction becomes nonzero when biological information helps reduce the task loss; otherwise, it can remain close to zero. This design prevents noisy biological knowledge from affecting routing at the beginning while allowing the model to use it when helpful.

More explicitly, $m$ indexes a token and $t$ indexes a recursion step; $h_m^{(t)} \in \mathbb{R}^d$ is row $m$ of $H^{(t)}$; $w_r \in \mathbb{R}^d$ is the base-router weight vector; and $\tau_r > 0$ is a temperature controlling how sharply router scores separate. The vector $[\,\cdot,\cdot\,]_m$ concatenates token $m$'s neighborhood mean, $[AH^{(t)}]_m$, with its neighborhood residual, $[H^{(t)} - AH^{(t)}]_m$. The small network $\phi_\psi$, with parameters $\psi$, maps this evidence to one scalar correction. The sigmoid $\sigma(\gamma_t) \in (0,1)$ is a learned step-specific gate. A higher corrected score makes another recursive update more likely under the chosen routing policy. Performance can improve when the correction spends depth on tokens whose local biological context is informative; if the graph is unhelpful, zero initialization and the learned gate provide a route back toward ordinary MoR behavior.

**Training and efficiency.** The main task loss is class-weighted cross-entropy. During training we minimize

$$\mathcal{L} = \mathcal{L}_{\mathrm{CE}} + \alpha\mathcal{L}_{\mathrm{suf}} + \beta\mathcal{L}_{\mathrm{div}} + \gamma\mathcal{L}_{\mathrm{cmp}} + \eta\mathcal{L}_{\mathrm{router}}, \quad (5)$$

where the auxiliary terms encourage marker sufficiency ($\mathcal{L}_{\mathrm{suf}}$), marker diversity ($\mathcal{L}_{\mathrm{div}}$), marker compression/sparsity ($\mathcal{L}_{\mathrm{cmp}}$), and stable router logits and balanced token-choice routing ($\mathcal{L}_{\mathrm{router}}$). $\mathcal{L}_{\mathrm{CE}}$ is the class-weighted prediction loss. The nonnegative coefficients $\alpha$, $\beta$, $\gamma$, and $\eta$ set the relative influence of the four auxiliary objectives. Increasing one coefficient emphasizes its behavior but may reduce attention to the primary prediction objective, so any selection of these coefficients must use validation data rather than test performance. They should not be interpreted as independent performance gains. Sufficiency discourages selecting markers that lose predictive signal; diversity discourages redundant selections; compression encourages a compact set; and the router term stabilizes token-choice utilization.

In expert-choice routing, each recursion step selects a fixed number of the highest-scoring tokens. This fixed capacity directly controls how many tokens are processed at each step, so an additional load-balancing loss is not needed (Bae et al. 2025). To quantify computational cost, we report FLOPs, a hardware-independent count of the arithmetic operations performed by the Transformer stack:

$$\Phi_{\mathrm{eff}} = \sum_{t=1}^{K} \left(4a_t^2 d + 4a_t d d_{\mathrm{ff}}\right), \quad (6)$$

where $a_t$ is the mean number of active tokens at recursion step $t$, $d$ is the model width, and $d_{\mathrm{ff}}$ is the hidden dimension of the feed-forward network. The first term grows quadratically with the active-token count and represents attention-like token interactions, while the second term represents feed-forward work and grows linearly with $a_t$. $\Phi_{\mathrm{eff}}$ sums these costs over the executed recursion steps. Routing reduces computation when $a_t < M$ at later steps; because the same $f_\theta$ is reused, this compute saving is separate from the parameter saving obtained through weight sharing. The practical trade-off is controlled by $K$ and the routing capacity: more active tokens or deeper recursion can improve refinement, but increases FLOPs and may provide diminishing returns.

# Appendix E: Extended Results

| | | | | | Single-cell macro-F1 ↑ | | | | | | | | Multi-omics macro-F1 ↑ | | | | |
|---|---|---|---|---|---|---|---|---|---|---|---|---|---|---|---|---|---|
| Model | Type | $N_R$ | Param | FLOPs | Segerst. | Lung | Oesoph. | Baron | Muraro | T-cell | Spleen | Xin | BLCA | STAD | PAN-2M | PAN-3M | Avg |
| Vanilla | – | 4 | 300K | 1.00× | 62.5±11.2 | 72.5±2.5 | 58.7±4.2 | 61.0±8.5 | 80.2±4.3 | 52.1±3.5 | 48.3±3.7 | 74.0±12.1 | 40.6±5.8 | 44.5±7.5 | 75.6±3.5 | 95.2±1.5 | 66.1±5.3 |
| Recursive | – | 2 | 75K | 0.50× | 65.8±4.4 | 71.7±2.8 | 57.3±2.9 | 65.6±2.8 | 79.8±6.5 | 53.0±2.0 | 48.8±2.3 | 68.2±2.4 | 41.4±1.3 | 46.0±3.4 | 75.7±2.9 | 94.9±1.5 | 66.3±3.5 |
| | – | 3 | 75K | 0.75× | 64.5±3.1 | 72.7±2.0 | 57.5±2.1 | 67.1±8.7 | 78.7±4.9 | 49.8±4.4 | 49.4±3.3 | 67.8±7.3 | 43.2±6.2 | 45.7±3.7 | 74.6±2.7 | 95.0±1.4 | 66.0±4.1 |
| | – | 4 | 75K | 1.00× | 61.4±7.3 | 72.9±2.9 | 56.9±4.3 | 65.9±2.7 | 75.8±4.1 | 50.5±3.5 | 49.4±3.4 | 72.2±5.2 | 40.2±1.1 | 43.2±6.0 | 72.3±1.8 | 94.3±1.1 | 65.5±3.7 |
| **MoR** | Expert | 2 | 75K | 0.42× | 65.0±6.2 | 72.9±1.7 | 57.0±3.5 | 58.0±6.1 | 72.3±2.9 | 50.6±4.9 | 51.7±1.4 | 68.0±8.3 | 39.7±12.3 | 49.7±6.3 | 77.7±1.4 | 94.6±1.3 | 65.9±4.7 |
| | Expert | 3 | 75K | 0.59× | 60.7±6.9 | 74.2±1.7 | 58.4±3.6 | 61.6±2.3 | 69.3±7.2 | 52.3±2.4 | 47.5±3.0 | 65.7±7.5 | 38.0±8.0 | 42.3±4.4 | 77.0±1.1 | 94.6±1.6 | 64.5±4.3 |
| | Expert | 4 | 75K | 0.76× | 64.3±11.6 | 72.3±2.7 | 55.3±2.6 | 68.9±3.7 | 74.8±4.7 | 50.9±4.7 | 53.2±1.3 | 66.1±9.7 | 43.7±2.8 | 45.2±4.1 | 75.5±3.0 | 94.3±1.3 | 65.8±4.4 |
| | Token | 2 | 75K | 0.42× | 57.4±6.6 | 72.2±2.7 | 57.7±1.8 | 66.2±8.5 | 82.5±7.1 | 50.7±2.2 | 50.6±2.5 | 68.9±6.0 | 35.7±7.8 | 47.7±4.0 | 76.4±4.6 | 94.3±2.8 | 66.6* |
| | Token | 3 | 75K | 0.52× | 55.1±5.0 | 70.5±3.3 | 56.6±3.4 | 64.0±4.6 | 75.9±4.7 | 55.8±3.0 | 49.2±2.1 | 70.1±3.5 | 41.3±8.6 | 46.5±4.5 | 77.2±3.9 | 93.6±2.1 | 65.6* |
| | Token | 4 | 75K | 0.56× | 60.7±5.8 | 73.3±2.4 | 59.8±3.7 | 65.2±5.5 | 81.6±2.7 | 50.4±2.2 | 51.1±3.1 | 62.3±5.7 | 40.6±6.9 | 48.8±3.6 | 78.0±4.6 | 95.3±1.6 | 67.0±3.6 |
| **bioMoR (ours)** | Expert | 2 | 75K | 0.42× | 72.4±2.9 | **80.3±1.5** | 71.0±1.5 | **85.2±2.9** | 83.8±3.1 | 69.2±0.7 | 60.3±1.9 | 76.3±9.6 | 28.2±4.5 | 43.6±3.9 | 85.5±1.9 | 96.0±1.3 | 71.4±2.8 |
| | Expert | 3 | 75K | 0.59× | **76.1±2.9** | 80.1±0.3 | 69.2±1.6 | 81.6±2.7 | **85.1±7.6** | 69.6±0.6 | **61.4±1.7** | **80.1±7.0** | 34.2±6.5 | 49.2±5.0 | 86.0±1.0 | 96.3±0.8 | 72.7±3.1 |
| | Expert | 4 | 75K | 0.76× | 72.2±4.5 | 80.0±1.2 | 70.8±1.6 | 81.9±1.9 | 83.7±4.9 | **70.0±1.9** | 59.7±0.8 | 72.7±5.5 | **47.1±7.1** | 46.0±2.6 | 85.8±1.5 | **98.0±0.4** | 72.6±2.8 |
| **+ Token choice** | Token | 2 | 75K | 0.42× | 74.4±6.8 | 79.4±1.3 | 71.1±2.4 | 82.9±5.0 | 83.8±3.2 | 68.6±1.7 | 60.3±1.2 | 69.1±6.6 | 45.0±13.8 | 44.2±3.5 | 85.1±1.5 | 95.5±2.2 | 71.6±1.6 |
| | Token | 3 | 75K | 0.52× | 73.6±7.0 | 80.2±1.5 | 71.0±1.2 | 83.9±3.0 | 78.4±4.8 | 68.6±0.9 | 61.0±1.0 | 77.2±5.3 | 39.6±13.7 | 45.4±4.8 | 84.8±2.3 | 95.8±1.0 | 71.6±1.5 |
| | Token | 4 | 75K | 0.56× | 75.8±2.8 | 79.7±0.9 | 70.8±2.1 | 82.9±4.0 | 81.9±6.8 | 69.8±1.2 | 61.0±0.6 | 76.1±8.9 | 39.5±5.9 | **51.4±5.0** | **86.0±1.9** | 96.4±1.6 | **73.0±3.0** |

Table 3: **Complete architecture ladder (5-fold CV).** Macro-F1 (mean±SD) for all 11 classification datasets. Parameters and Transformer-stack FLOPs are reported relative to the four-layer Vanilla Transformer. PM, PC, and 3M denote pan-cancer mutation+CNV, expression, and tri-modal inputs, respectively. This expands the representative eight-dataset table in the main paper; bold, underlined values mark the best result in each dataset column.

| | | | | | Single-cell balanced acc. ↑ | | | | | | | | Multi-omics balanced acc. ↑ | | | | |
|---|---|---|---|---|---|---|---|---|---|---|---|---|---|---|---|---|---|
| Model | Type | $N_R$ | Param | FLOPs | Segerst. | Lung | Oesoph. | Baron | Muraro | T-cell | Spleen | Xin | BLCA | STAD | PAN-2M | PAN-3M | Avg |
| Vanilla | – | 4 | 300K | 1.00× | 71.7±10.3 | 75.2±1.5 | 60.0±2.9 | 67.7±2.6 | 79.1±3.3 | 55.2±5.7 | 55.3±1.4 | 73.6±8.7 | 48.3±2.8 | 48.4±3.8 | 74.1±4.0 | 94.9±1.9 | 67.1* |
| Recursive | – | 2 | 75K | 0.50× | 69.2±11.0 | 75.9±1.5 | 61.1±4.1 | 71.0±5.6 | 77.1±4.6 | 57.4±6.1 | 57.8±1.5 | 73.1±9.7 | 48.9±5.4 | 49.3±3.9 | 72.7±4.2 | 94.7±2.1 | 67.1* |
| | – | 3 | 75K | 0.75× | 74.9±9.7 | 75.0±1.7 | 61.1±3.2 | 71.1±3.4 | 79.8±5.4 | 55.7±4.7 | 57.9±1.8 | 74.1±9.8 | 46.0±3.7 | 50.2±2.5 | 71.7±5.2 | 95.5±2.8 | 67.5* |
| | – | 4 | 75K | 1.00× | 69.6±9.1 | 75.3±1.9 | 60.2±3.6 | 70.8±6.9 | 77.8±2.4 | 56.5±4.6 | 56.9±1.5 | 76.5±9.4 | 48.5±2.8 | 49.6±3.5 | 74.4±1.6 | 93.6±1.7 | 67.2* |
| **MoR** | Expert | 2 | 75K | 0.42× | 65.9±3.6 | 75.6±1.3 | 61.2±3.0 | 68.7±3.8 | 76.7±5.0 | 58.1±2.3 | 57.8±1.3 | 75.2±5.8 | 46.5±3.0 | 51.4±4.1 | 76.4±3.1 | 94.8±1.5 | 67.7* |
| | Expert | 3 | 75K | 0.59× | 70.9±4.3 | 76.2±1.4 | 61.4±3.2 | 68.4±2.0 | 81.4±5.8 | 56.0±5.5 | 58.0±1.5 | 75.3±6.7 | 47.1±4.5 | 49.1±2.6 | 74.6±5.3 | 92.5±3.1 | 67.3* |
| | Expert | 4 | 75K | 0.76× | 69.2±6.1 | 76.5±1.8 | 62.7±3.7 | 70.9±2.7 | 75.5±2.3 | 56.6±6.1 | 57.8±2.5 | 74.1±10.0 | 49.8±5.0 | 47.0±1.3 | 74.6±3.1 | 93.7±3.1 | 67.8* |
| | Token | 2 | 75K | 0.42× | 62.6±5.1 | 76.6±1.9 | 61.5±2.1 | 65.2±1.8 | 74.4±6.7 | 57.6±4.1 | 55.6±1.5 | 65.5±9.3 | 45.2±2.8 | 52.1±5.8 | 77.7±2.2 | 93.3±1.9 | 66.4* |
| | Token | 3 | 75K | 0.52× | 67.3±11.7 | 77.8±1.5 | 63.9±2.6 | 69.9±6.5 | 77.5±5.8 | 52.4±5.2 | 56.5±2.3 | 68.6±9.8 | 47.3±3.1 | 49.9±4.7 | 74.5±6.0 | 94.6±1.9 | 67.2* |
| | Token | 4 | 75K | 0.56× | 66.4±8.0 | 76.1±2.2 | 61.4±2.3 | 67.3±4.5 | 80.6±8.5 | 56.6±4.1 | 56.5±1.4 | 72.9±10.9 | 46.7±7.8 | 47.4±3.3 | 73.5±2.2 | 93.7±4.6 | 66.7* |
| **bioMoR (ours)** | Expert | 2 | 75K | 0.42× | 73.1±3.6 | 81.5±1.0 | 72.0±1.7 | 82.2±3.1 | 84.9±6.2 | 70.2±1.5 | **65.4±1.5** | 75.0±8.3 | 48.9±3.3 | **56.1±6.3** | 83.4±1.1 | **96.4±1.2** | 73.3* |
| | Expert | 3 | 75K | 0.59× | 77.5±4.8 | 80.7±1.6 | 71.2±0.3 | **87.4±4.8** | 82.5±8.7 | 70.8±1.5 | 64.3±0.9 | 78.3±6.7 | 49.0±0.8 | 50.2±1.6 | 85.5±1.5 | 95.6±2.5 | 73.5* |
| | Expert | 4 | 75K | 0.76× | 78.8±4.9 | 80.4±0.6 | 72.1±1.6 | 85.8±2.7 | **88.9±4.0** | 70.4±1.6 | 64.7±1.6 | 79.7±5.2 | 48.7±7.5 | 51.0±4.0 | 82.9±3.3 | 95.4±1.3 | **73.9*** |
| **+ Token choice** | Token | 4 | 75K | 0.56× | 73.3±7.0 | **82.0±1.5** | **73.2±1.6** | 86.4±3.8 | 83.4±3.7 | **71.3±1.4** | 64.1±1.1 | 81.7±5.8 | 48.3±2.2 | 46.7±3.0 | 83.4±2.6 | 96.1±1.9 | 73.3* |

Table 4: **Balanced accuracy across the classification suite (5-fold CV).** Each populated cell reports mean±SD balanced accuracy over the same five folds and architecture ladder used for macro-F1. Bold, underlined values mark the best result in each dataset column; starred averages exclude unavailable runs from the corresponding architecture configuration.

## Appendix F: Statistical Significance

We use the paired Wilcoxon signed-rank test introduced by Wilcoxon (1945) and follow the multiple-dataset classifier-comparison guidance of Demšar (2006).

Table 5: **Per-dataset macro-F1 (5-fold CV) and bioMoR's improvement over the strongest baseline.** Each column is one dataset; bold marks the best model. $\Delta$ vs best is bioMoR minus the per-dataset maximum of {Vanilla, Recursive, MoR}. bioMoR wins 11/12 (one-sided Wilcoxon signed-rank $p$=0.0007; vs. Recursive and vs. MoR individually 12/12, $p$=0.0002).

| Model | Seger. | Lung | Oeso. | Baron | Muraro | T-cell | Spleen | Xin | BLCA | STAD | PAN-2M | PAN-3M |
|---|---|---|---|---|---|---|---|---|---|---|---|---|
| Vanilla | 62.5 | 72.5 | 58.7 | 65.7 | 80.2 | 52.1 | 48.3 | **74.0** | 40.6 | 44.5 | 75.6 | 95.2 |
| Recursive | 61.4 | 72.9 | 56.9 | 70.3 | 75.8 | 50.5 | 49.4 | 72.2 | 40.2 | 43.2 | 72.3 | 94.3 |
| MoR | 64.3 | 72.3 | 55.3 | 68.9 | 74.8 | 50.9 | 53.2 | 66.1 | 43.7 | 45.2 | 75.5 | 94.3 |
| **bioMoR** | **72.2** | **80.0** | **70.8** | **81.9** | **83.7** | **70.0** | **59.7** | 72.7 | **47.1** | **46.0** | **85.8** | **98.0** |
| $\Delta$ vs best | +7.9 | +7.1 | +12.1 | +11.6 | +3.5 | +17.9 | +6.5 | -1.3 | +3.4 | +0.8 | +10.2 | +2.8 |
| Wilcoxon $p$ | $p = 0.0007$ (one-sided, bioMoR $>$ best baseline, $n = 12$) | | | | | | | | | | | |
| Win | ✓ | ✓ | ✓ | ✓ | ✓ | ✓ | ✓ | × | ✓ | ✓ | ✓ | ✓ |

Table 6: **Per-dataset balanced accuracy (5-fold CV) and bioMoR's improvement over the strongest baseline.** Layout as in Table 5. bioMoR wins 11/12 (one-sided Wilcoxon signed-rank $p$=0.0007; vs. Vanilla and vs. Recursive individually 12/12, $p$=0.0002).

| Model | Seger. | Lung | Oeso. | Baron | Muraro | T-cell | Spleen | Xin | BLCA | STAD | PAN-2M | PAN-3M |
|---|---|---|---|---|---|---|---|---|---|---|---|---|
| Vanilla | 71.7 | 75.2 | 60.0 | 67.7 | 79.1 | 55.2 | 55.3 | 73.6 | 48.3 | 48.4 | 74.1 | 94.9 |
| Recursive | 69.6 | 75.3 | 60.2 | 70.8 | 77.8 | 56.5 | 56.9 | 76.5 | 48.5 | 49.6 | 74.4 | 93.6 |
| MoR | 69.2 | 76.5 | 62.7 | 70.9 | 75.5 | 56.6 | 57.8 | 74.1 | **49.8** | 47.0 | 74.6 | 93.7 |
| **bioMoR** | **78.8** | **80.4** | **72.1** | **85.8** | **88.9** | **70.4** | **64.7** | **79.7** | 48.7 | **51.0** | **82.9** | **95.4** |
| $\Delta$ vs best | +7.1 | +3.9 | +9.4 | +15.0 | +9.8 | +13.8 | +6.9 | +3.2 | -1.1 | +1.4 | +8.3 | +0.5 |
| Wilcoxon $p$ | $p = 0.0007$ (one-sided, bioMoR $>$ best baseline, $n = 12$) | | | | | | | | | | | |
| Win | ✓ | ✓ | ✓ | ✓ | ✓ | ✓ | ✓ | ✓ | × | ✓ | ✓ | ✓ |

## Appendix G: Extended Survival Prediction

To test whether bioMoR transfers beyond classification, we evaluate survival-risk prediction on the PAN-2M pan-cancer cohort and four cancer-specific TCGA cohorts: KIRP, COAD, UCEC, and HNSC. The comparison includes bioMoR-Expert and bioMoR-Token together with the Vanilla, Recursive, MoR-Expert, and MoR-Token baselines. Performance is measured using Harrell's C-index; a larger value indicates better concordance between the predicted risk ranking and observed survival outcomes.

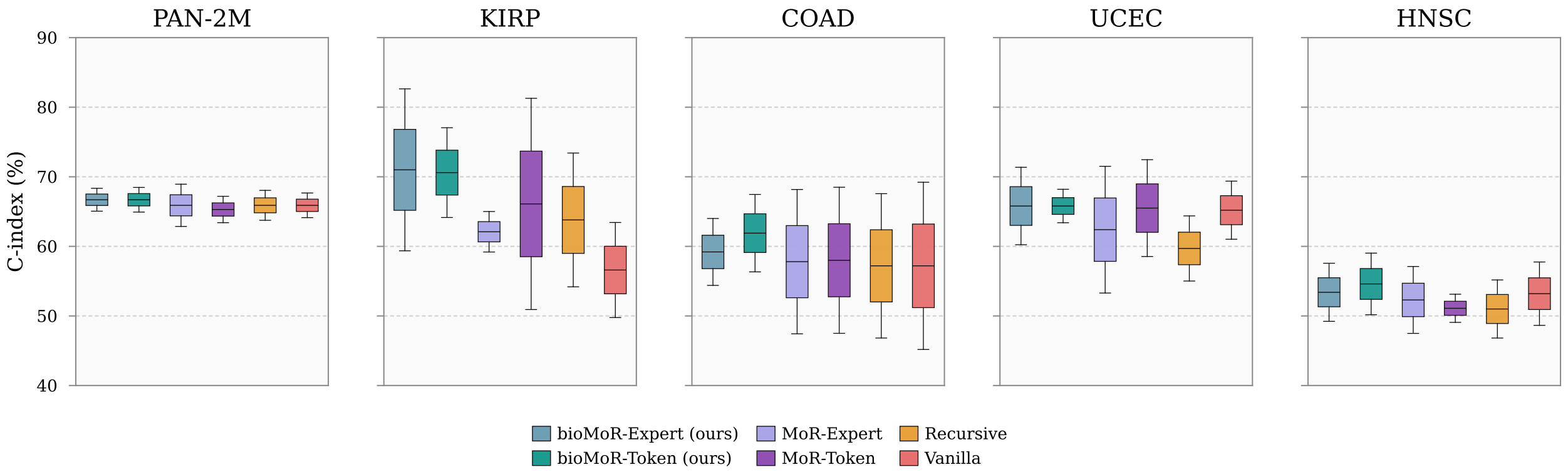


Figure 1: **Survival prediction across pan-cancer and TCGA cohorts.** Fold-level C-index distributions for PAN-2M, KIRP, COAD, UCEC, and HNSC. The bioMoR variants remain competitive across the pan-cancer and cancer-specific settings, supporting the transfer of biology-guided adaptive recursion to a time-to-event endpoint. Higher values indicate better concordance.

# Appendix H: Embedding Visualization with UMAP

To examine whether predictive performance is reflected in the learned representation, we freeze each model's penultimate embedding, fit a logistic-regression probe on the training samples, and evaluate it on held-out samples. Each dataset panel compares Vanilla, Recursive, MoR, and bioMoR under the same protocol.

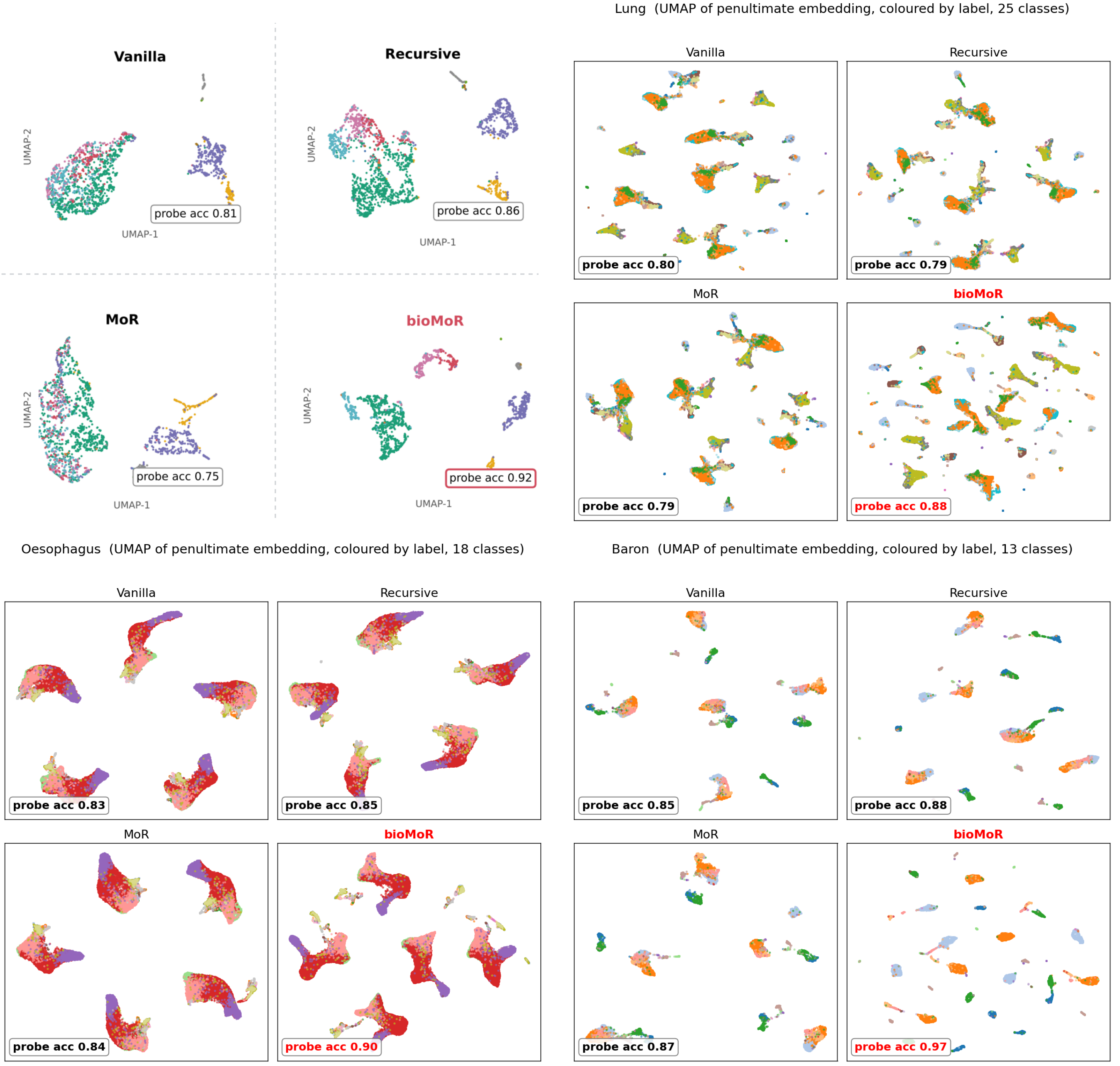


Figure 2: **Frozen-embedding UMAP comparisons for single-cell datasets (part 1).** Each dataset-level panel shows the four architecture variants, colors points by the ground-truth class, and reports held-out linear-probe accuracy within each subpanel. The Segerstolpe panel is the same figure used in the main paper.

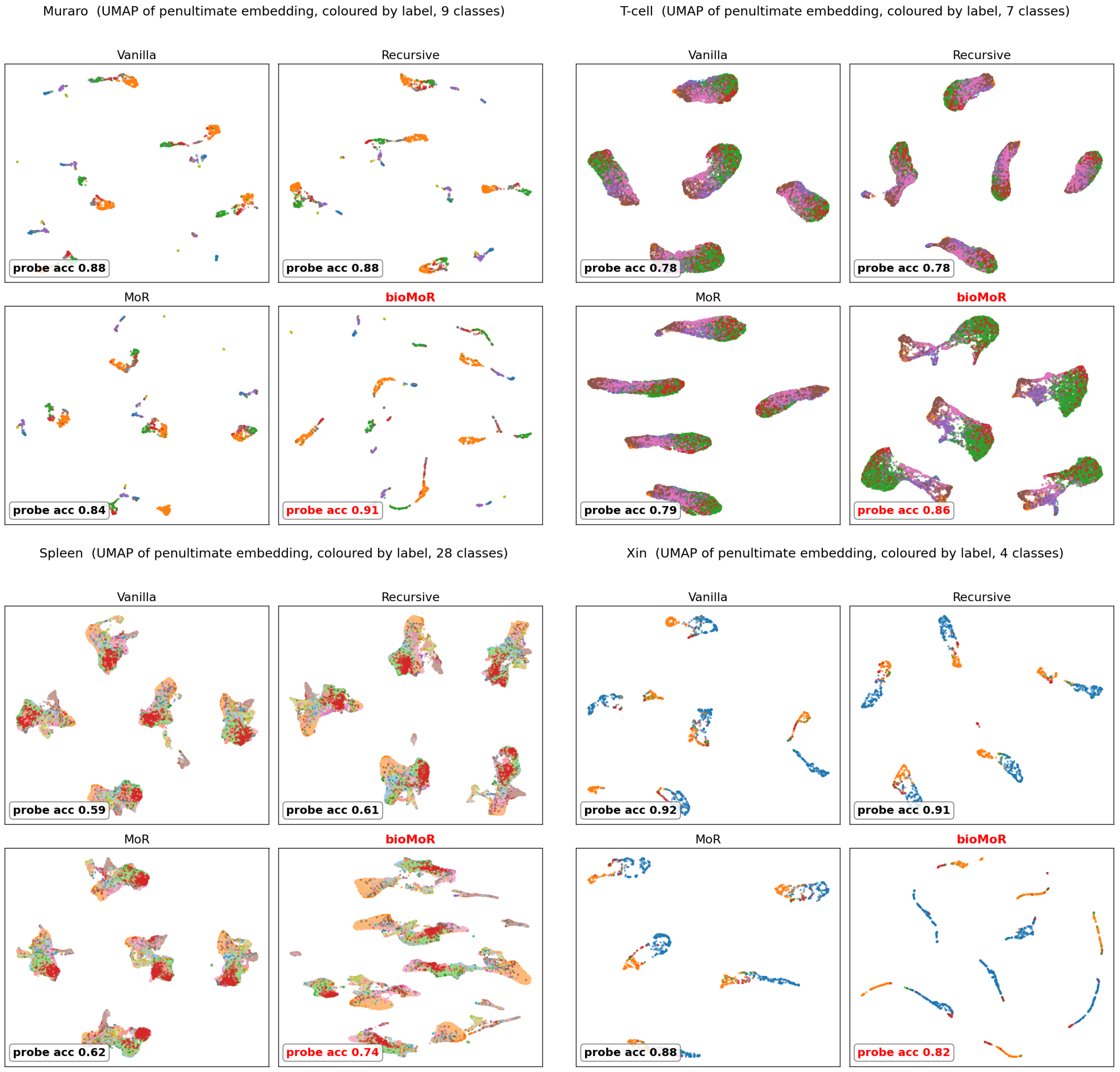


Figure 3: **Frozen-embedding UMAP comparisons for single-cell datasets (part 2).** The plotting, coloring, and held-out linear-probe protocol match Figure 2.

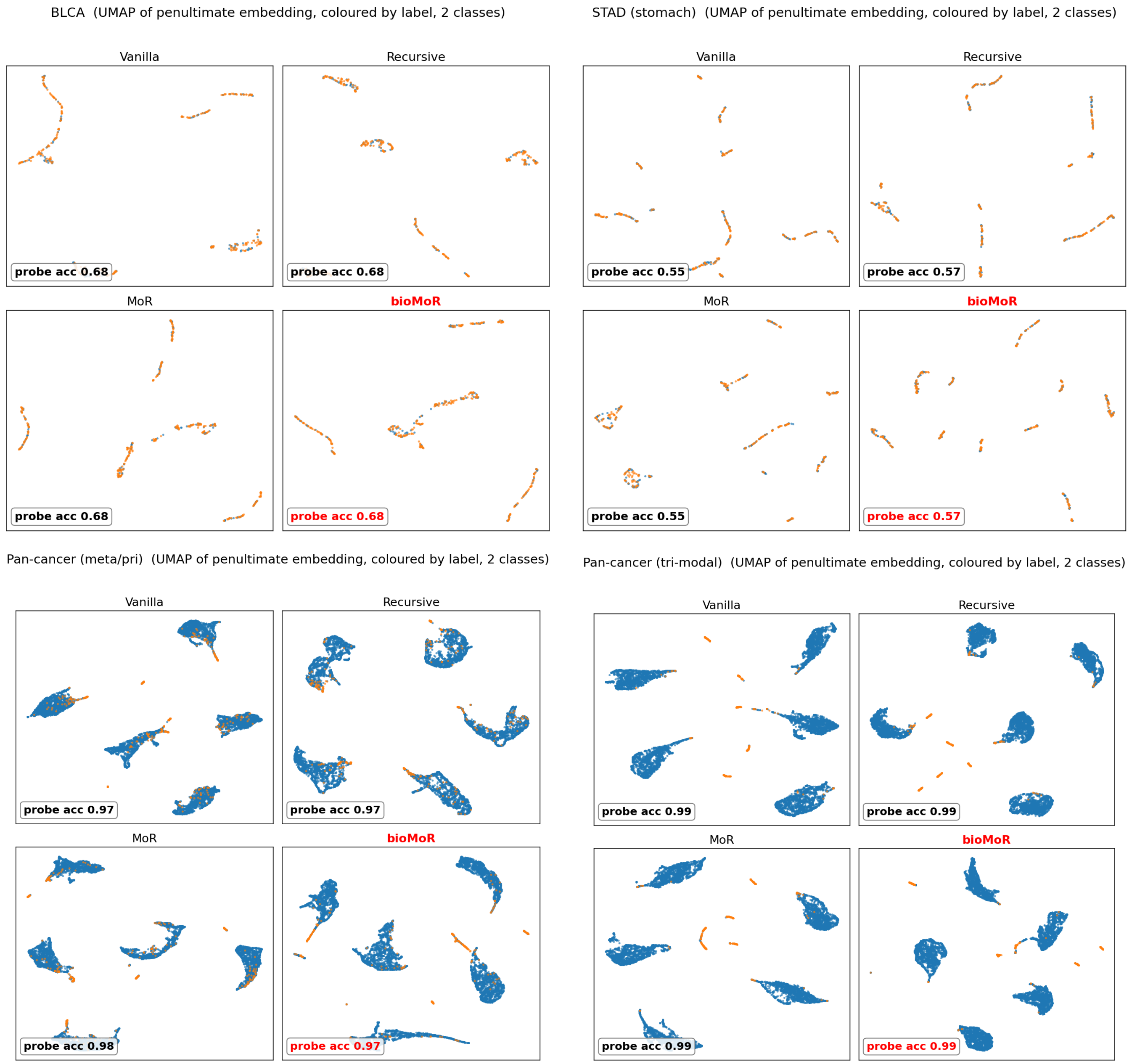


Figure 4: **Frozen-embedding UMAP comparisons for multi-omics datasets.** The plotting, coloring, and held-out linear-probe protocol match Figure 2; PAN-2M and PAN-3M denote the bi-modal and tri-modal pan-cancer cohorts, respectively.

## Appendix I: Ablation Studies

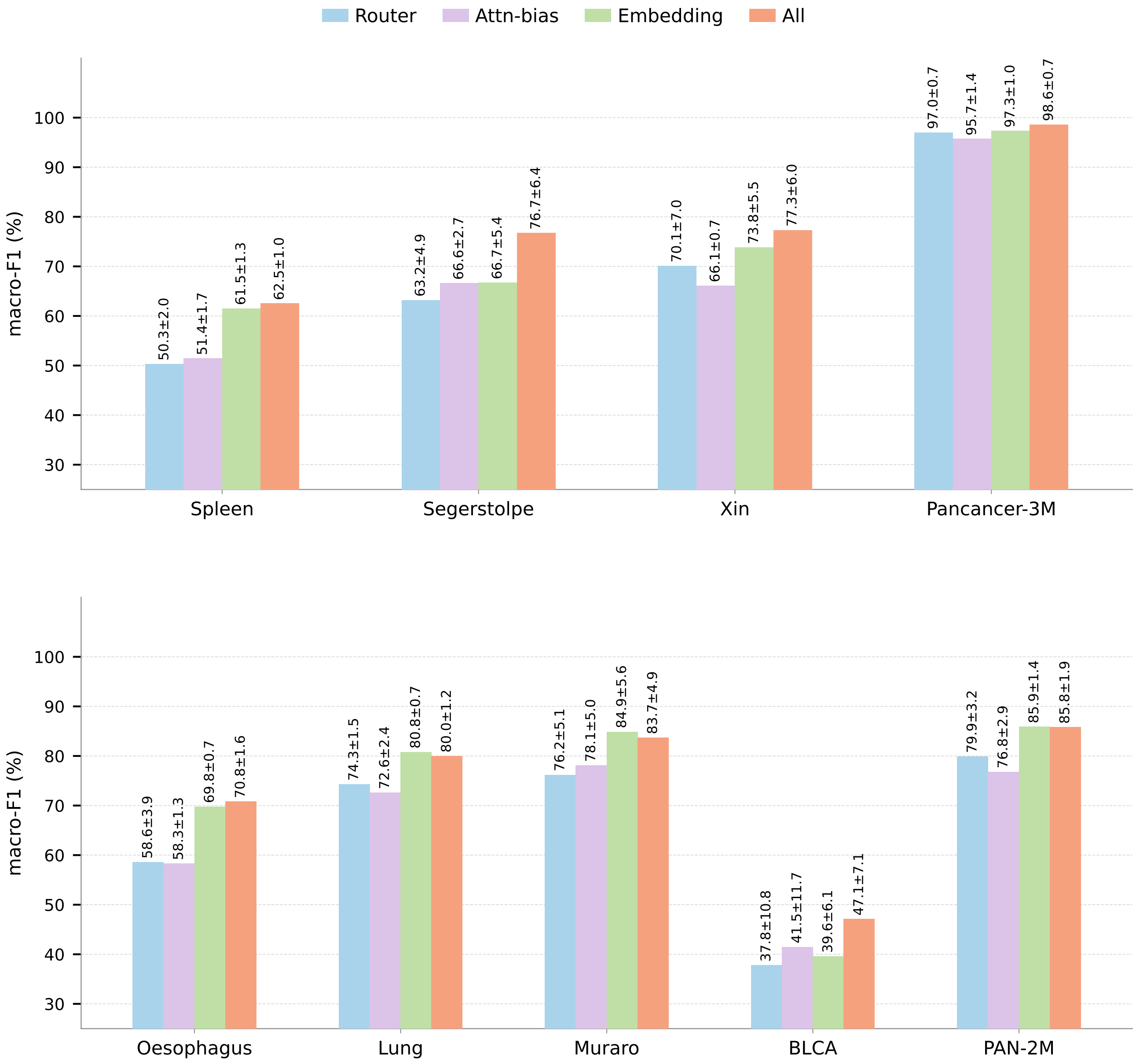


Figure 5: **Biological-injection ablation at recursion depth $K = 4$.** Macro-F1 (mean with fold-level variability) on the same four representative datasets shown in the main paper—Spleen, Segerstolpe, Xin, and Pancancer-3M— when biological information is applied only through the router, attention bias, or embedding, compared with the complete expert-choice bioMoR configuration. The first row matches the main-paper dataset order, labels, and values exactly; the second row extends the comparison to Oesophagus, Lung, Muraro, BLCA, and PAN-2M.

# Appendix J: Training and Validation Dynamics

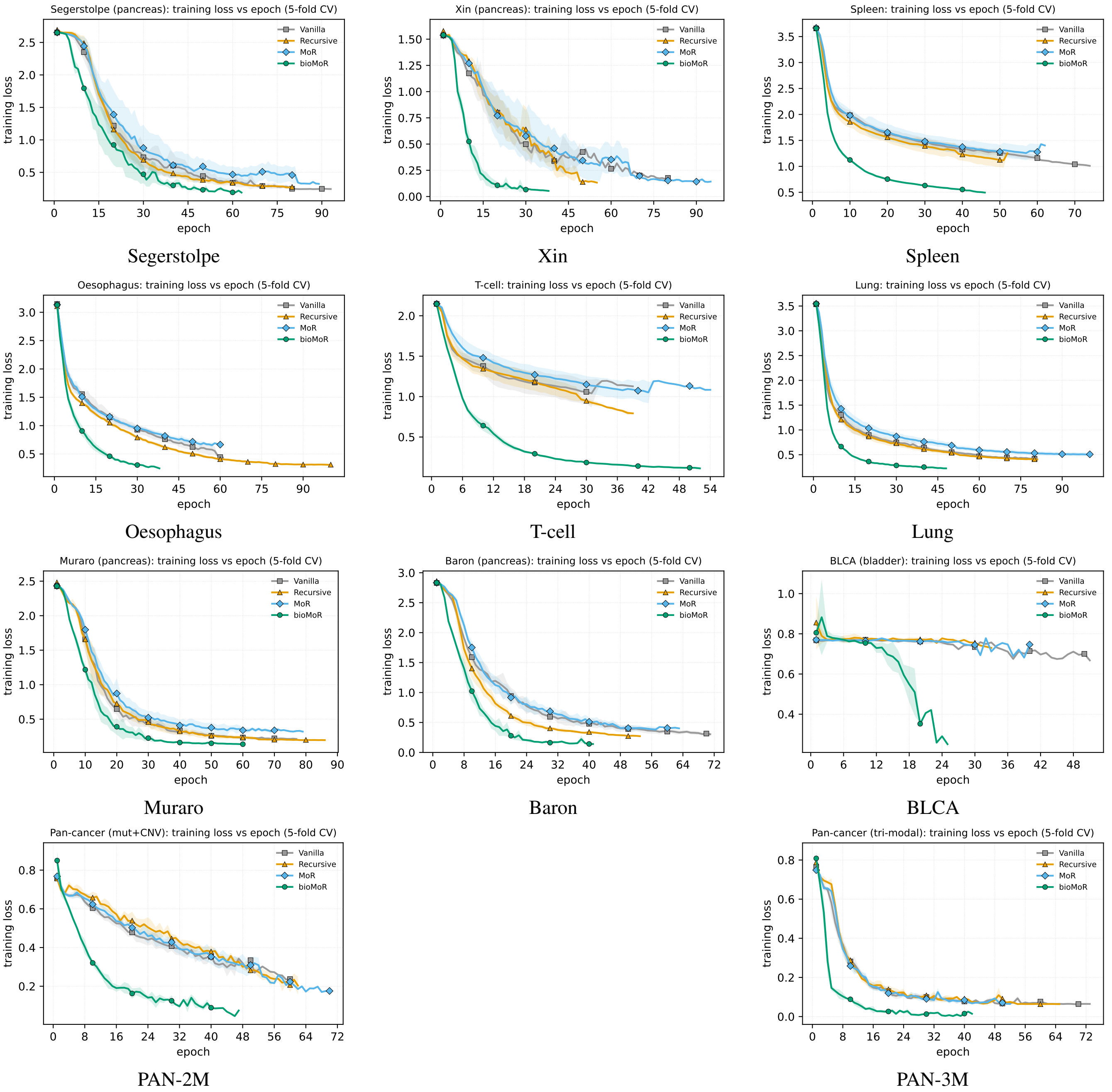


Figure 6: **Training loss across epochs.** Fold-aggregated training-loss trajectories for the available single-cell and multi-omics datasets used in the RQ3 convergence analysis.

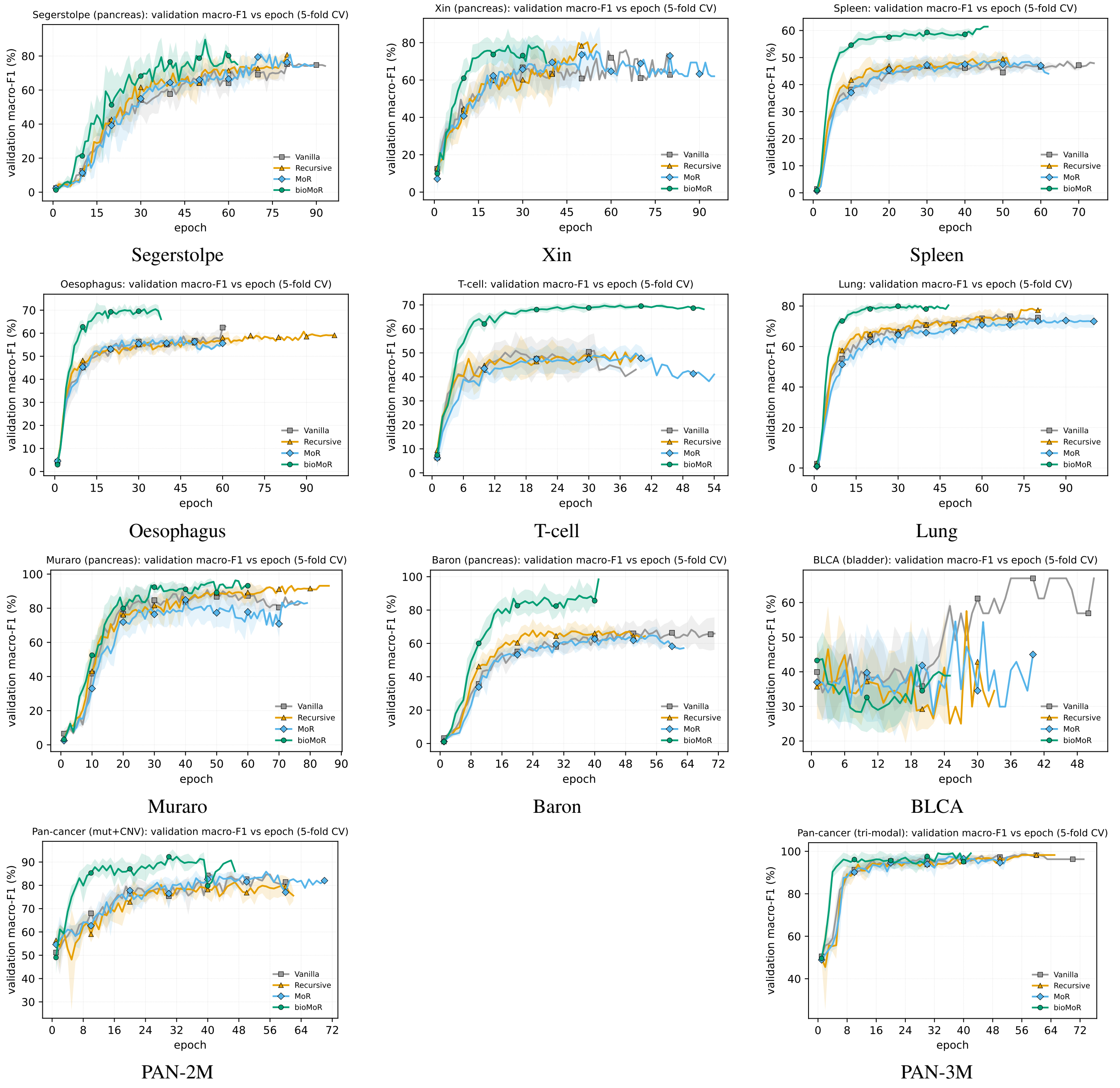


Figure 7: **Validation macro-F1 across epochs.** Fold-aggregated validation macro-F1 trajectories for all datasets with available RQ3 validation curves. Higher values indicate better validation performance.

# Appendix K: Accuracy-Efficiency Trade-offs

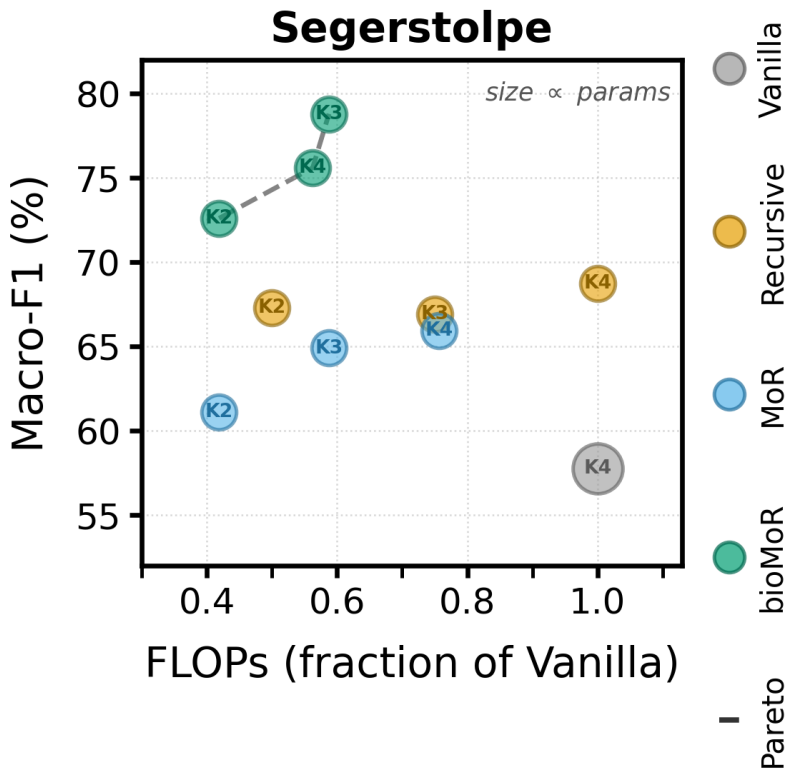


Segerstolpe

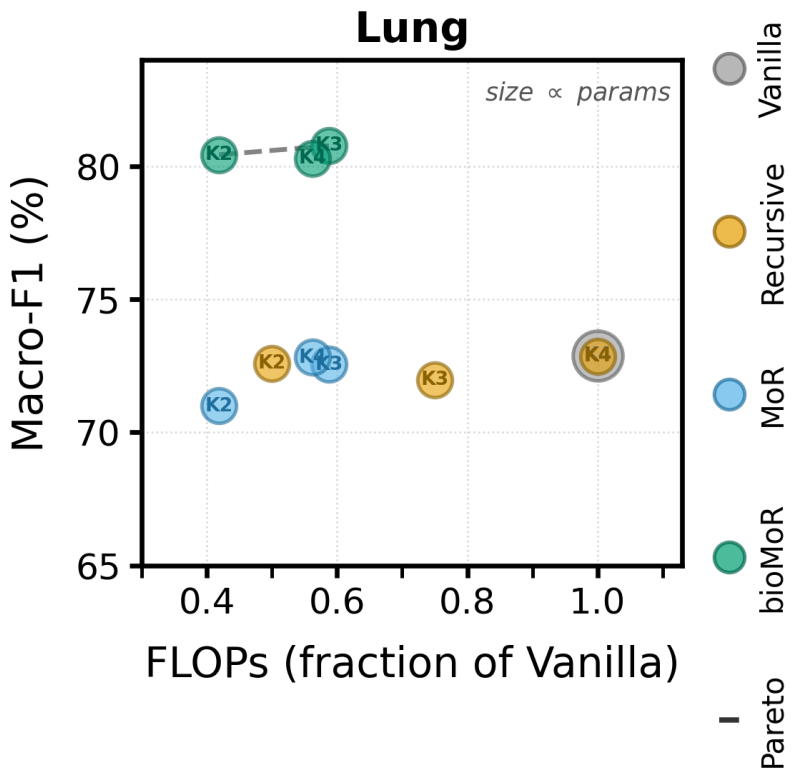


Lung

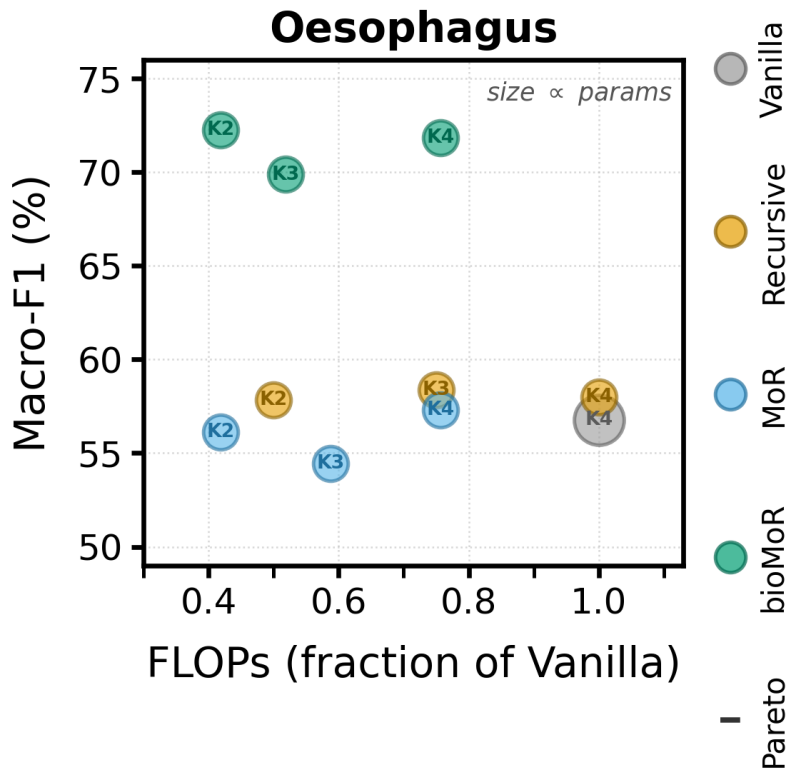


Oesophagus

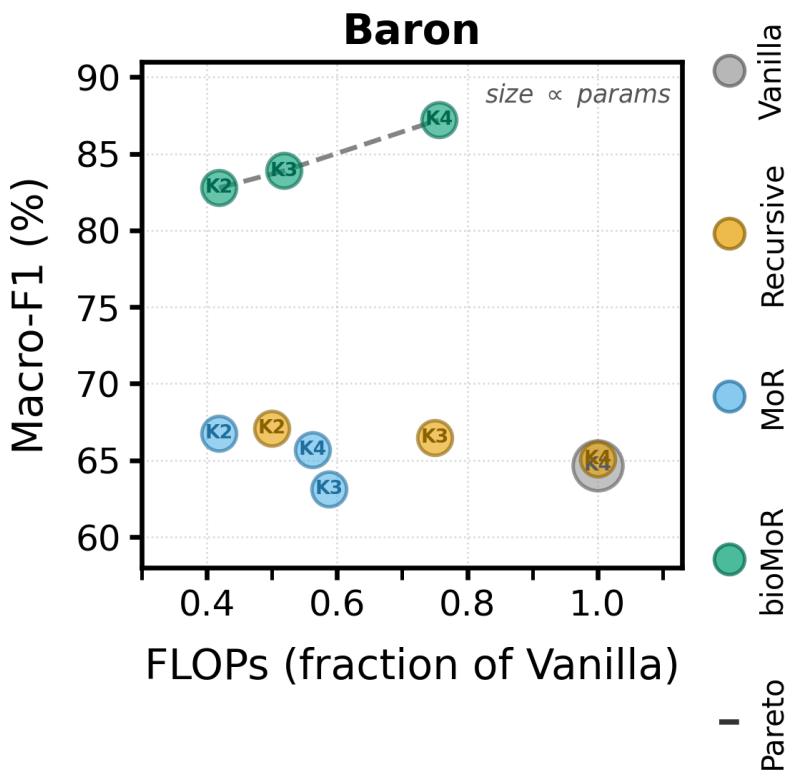


Baron

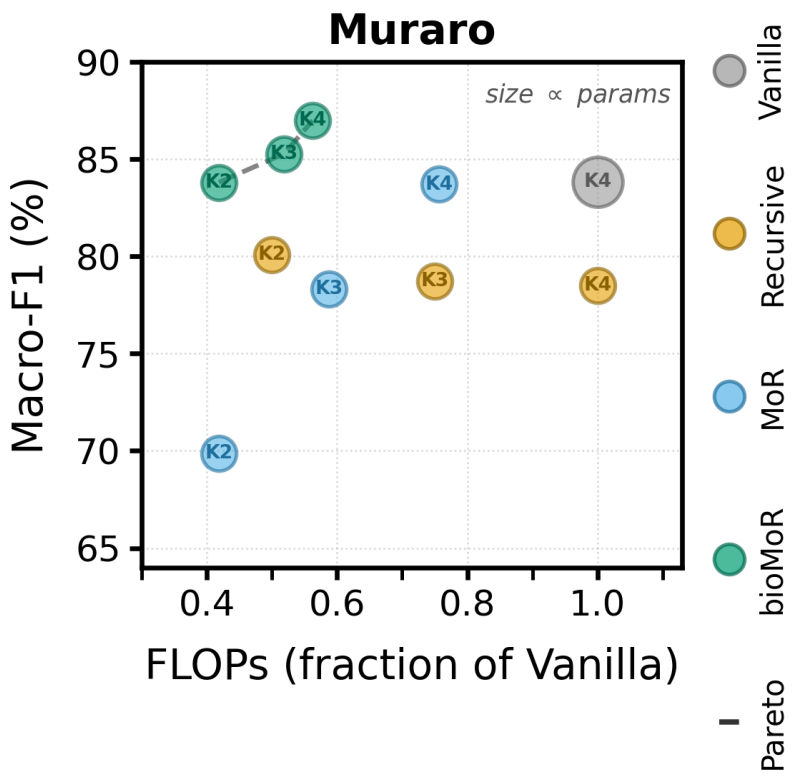


Muraro

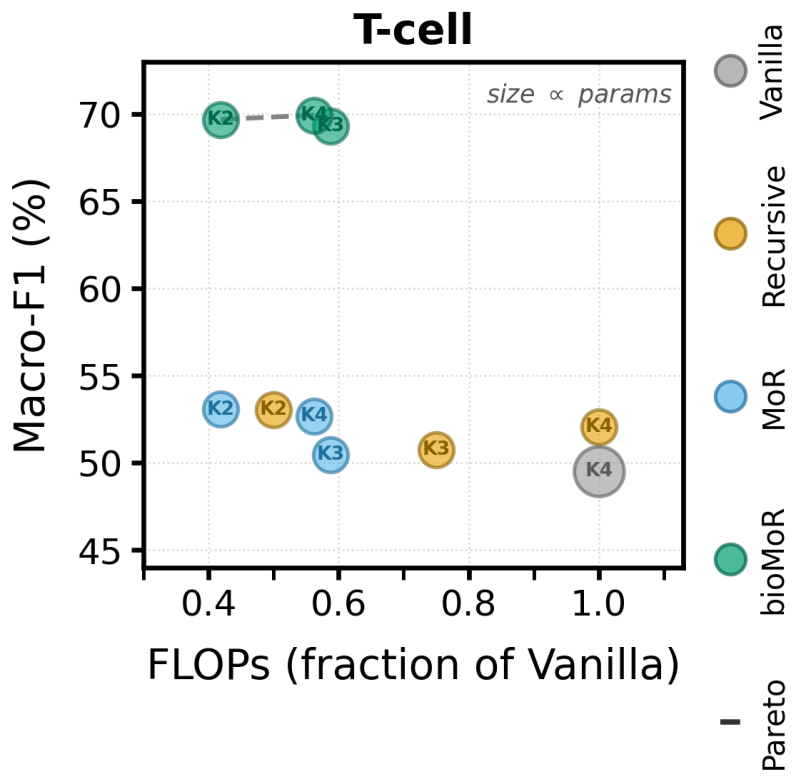


T-cell

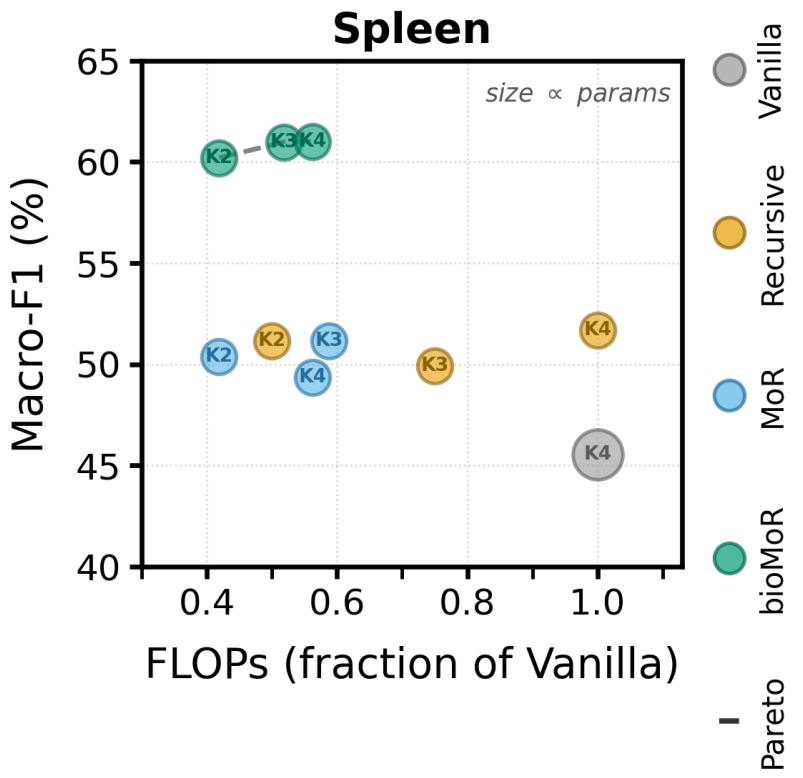


Spleen

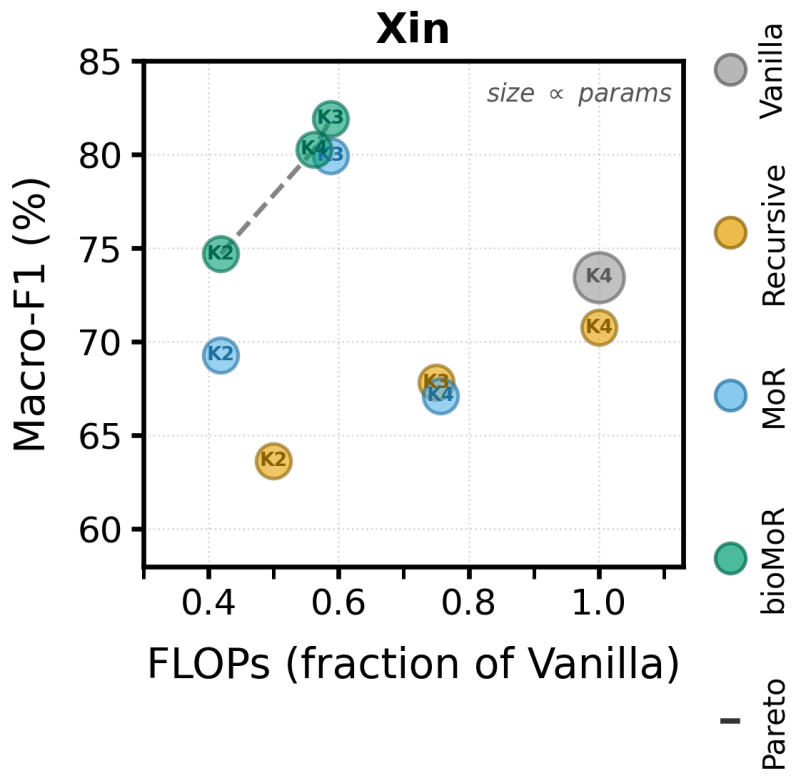


Xin

Figure 8: **Validation macro-F1 versus effective computation (part 1).** Dataset-level Pareto views compare the accuracy–efficiency trade-off across architectures and recursion settings; points toward the upper-left are preferred.

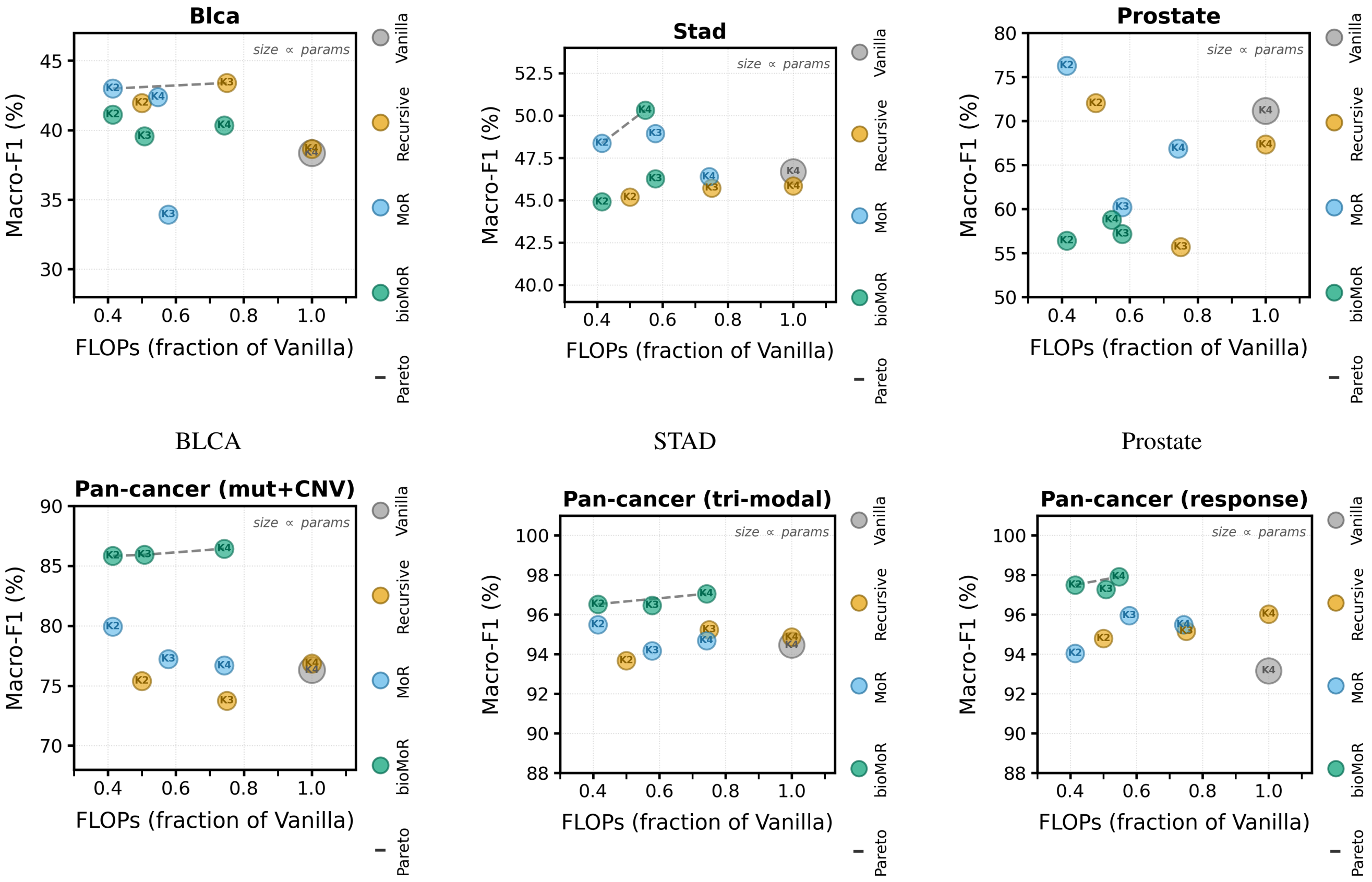


BLCA STAD Prostate

PAN-2M PAN-3M Pan-cancer response

Figure 9: **Validation macro-F1 versus effective computation (part 2).** Multi-omics Pareto views use the same axes and interpretation as Figure 8.

# Appendix L: Case Study: PAN-3M

## Pathway Markers by Recursion Depth ($M$=1257, $K$=4, Macro-F1 96.9±1.4)

Per-depth keep/erase rankings for the tri-modal pan-cancer cohort (PAN-3M; mutation+CNV+expression, 8586 samples, 2 classes) under canonical bioMoR at $K$=4 expert-choice routing with capacity floor RMT_CAP_FLOOR= 0.5625 (schedule $c = (1.00, 0.75, 0.5625, 0.5625)$). The pathway view tokenises all 1257 Reactome pathways using the learned pathway graph. The expert-choice funnel keeps 56% of pathway tokens to full depth (707/1257): tokens drop at depth 2 (to 75%) and again at depth 3 (to 56%), then hold. Pathway-view Macro-F1 is 96.9. Mean recursion depth $d_m \in [1, 4]$ is averaged over the 5 held-out folds. For each depth $d \in \{2, 3, 4\}$ we list up to 25 *kept* markers still active at depth $d$ (rounded $d_m \geq d$, deepest first) and up to 25 *erased* markers that exit entering depth $d$ (rounded $d_m = d-1$). Depth 1 keeps all $M$ tokens.

### Depth 2.

**Kept — active at depth 2**

| # | Pathway | $d_m$ |
|---|---|---|
| 1 | Intra-Golgi and retrograde Golgi-to-ER traffic | 3.97 |
| 2 | Glycerophospholipid biosynthesis | 3.96 |
| 3 | Post-translational modification: synthesis of GPI-anchored proteins | 3.95 |
| 4 | Golgi-to-ER retrograde transport | 3.95 |
| 5 | Factors involved in megakaryocyte development and platelet production | 3.95 |
| 6 | Semaphorin interactions | 3.95 |
| 7 | RUNX1 regulates genes involved in megakaryocyte differentiation and p… | 3.95 |
| 8 | FOXO-mediated transcription | 3.94 |
| 9 | Phospholipid metabolism | 3.94 |
| 10 | Kinesins | 3.94 |
| 11 | Chaperonin-mediated protein folding | 3.93 |
| 12 | Deubiquitination | 3.92 |
| 13 | Class I MHC mediated antigen processing & presentation | 3.92 |
| 14 | Signaling by Interleukins | 3.92 |
| 15 | SLC-mediated transmembrane transport | 3.90 |
| 16 | Coagulation pathway | 3.90 |
| 17 | Synthesis of PA | 3.90 |
| 18 | Regulation of clotting cascade | 3.90 |
| 19 | Regulation of MITF-M-dependent genes involved in apoptosis | 3.90 |
| 20 | MHC class II antigen presentation | 3.90 |
| 21 | Cell surface interactions at the vascular wall | 3.89 |
| 22 | Cellular Senescence | 3.88 |
| 23 | NOTCH3 Intracellular Domain Regulates Transcription | 3.88 |
| 24 | Ca2+ pathway | 3.87 |
| 25 | Mitotic Prometaphase | 3.85 |

**Erased — exit entering depth 2**

| # | Pathway | $d_m$ |
|---|---|---|
| 1 | GABA synthesis, release, reuptake and degradation | 1.48 |
| 2 | APC/C:Cdh1 mediated degradation of Cdc20 and other APC/C:Cdh1 targete… | 1.47 |
| 3 | Sensory processing of sound by inner hair cells of the cochlea | 1.47 |
| 4 | Unblocking of NMDA receptors, glutamate binding and activation | 1.47 |
| 5 | MAP3K8 (TPL2)-dependent MAPK1/3 activation | 1.45 |
| 6 | SHC-mediated cascade:FGFR3 | 1.45 |
| 7 | Phosphorylated BMAL1:CLOCK (ARNTL:CLOCK) activates expression of core… | 1.43 |
| 8 | Degradation of CRY and PER proteins | 1.43 |
| 9 | Transcription of E2F targets under negative control by p107 (RBL1) an… | 1.41 |
| 10 | Activation of G protein gated Potassium channels | 1.41 |
| 11 | Fibronectin matrix formation | 1.40 |
| 12 | MyD88-independent TLR4 cascade | 1.39 |
| 13 | Disorders of Developmental Biology | 1.36 |
| 14 | Trafficking of AMPA receptors | 1.34 |
| 15 | E3 ubiquitin ligases ubiquitinate target proteins | 1.28 |
| 16 | Anchoring fibril formation | 1.22 |
| 17 | JNK (c-Jun kinases) phosphorylation and activation mediated by activ… | 1.21 |
| 18 | FRS-mediated FGFR3 signaling | 1.17 |
| 19 | Paracetamol ADME | 1.14 |
| 20 | Aspirin ADME | 1.10 |
| 21 | Drug ADME | 1.06 |

**Depth 3.**

**Kept — active at depth 3**

| # | Pathway | $d_m$ |
|---|---|---|
| 1 | Intra-Golgi and retrograde Golgi-to-ER traffic | 3.97 |
| 2 | Glycerophospholipid biosynthesis | 3.96 |
| 3 | Post-translational modification: synthesis of GPI-anchored proteins | 3.95 |
| 4 | Golgi-to-ER retrograde transport | 3.95 |
| 5 | Factors involved in megakaryocyte development and platelet production | 3.95 |
| 6 | Semaphorin interactions | 3.95 |
| 7 | RUNX1 regulates genes involved in megakaryocyte differentiation and p... | 3.95 |
| 8 | FOXO-mediated transcription | 3.94 |
| 9 | Phospholipid metabolism | 3.94 |
| 10 | Kinesins | 3.94 |
| 11 | Chaperonin-mediated protein folding | 3.93 |
| 12 | Deubiquitination | 3.92 |
| 13 | Class I MHC mediated antigen processing & presentation | 3.92 |
| 14 | Signaling by Interleukins | 3.92 |
| 15 | SLC-mediated transmembrane transport | 3.90 |
| 16 | Coagulation pathway | 3.90 |
| 17 | Synthesis of PA | 3.90 |
| 18 | Regulation of clotting cascade | 3.90 |
| 19 | Regulation of MITF-M-dependent genes involved in apoptosis | 3.90 |
| 20 | MHC class II antigen presentation | 3.90 |
| 21 | Cell surface interactions at the vascular wall | 3.89 |
| 22 | Cellular Senescence | 3.88 |
| 23 | NOTCH3 Intracellular Domain Regulates Transcription | 3.88 |
| 24 | Ca2+ pathway | 3.87 |
| 25 | Mitotic Prometaphase | 3.85 |

**Erased — exit entering depth 3**

| # | Pathway | $d_m$ |
|---|---|---|
| 1 | Formation of WDR5-containing histone-modifying complexes | 2.50 |
| 2 | COPII-mediated vesicle transport | 2.50 |
| 3 | TP53 Regulates Transcription of Genes Involved in Cytochrome C Release | 2.49 |
| 4 | HIV Life Cycle | 2.49 |
| 5 | Translation of Structural Proteins_9694635 | 2.49 |
| 6 | Netrin-1 signaling | 2.49 |
| 7 | Glucagon-type ligand receptors | 2.49 |
| 8 | Signaling by BMP | 2.49 |
| 9 | Notch-HLH transcription pathway | 2.49 |
| 10 | Impaired BRCA2 binding to PALB2 | 2.49 |
| 11 | Processive synthesis on the C-strand of the telomere | 2.49 |
| 12 | Nicotinate metabolism | 2.49 |
| 13 | Negative regulators of DDX58/IFIH1 signaling | 2.49 |
| 14 | Aberrant regulation of mitotic exit in cancer due to RB1 defects | 2.48 |
| 15 | Infection with Enterobacteria | 2.48 |
| 16 | ADORA2B mediated anti-inflammatory cytokines production | 2.48 |
| 17 | ABC-family protein mediated transport | 2.48 |
| 18 | Degradation of GLI1 by the proteasome | 2.48 |
| 19 | Striated Muscle Contraction | 2.47 |
| 20 | Proteasome assembly | 2.47 |
| 21 | Assembly of collagen fibrils and other multimeric structures | 2.47 |
| 22 | Sensory perception of sweet, bitter, and umami (glutamate) taste | 2.47 |
| 23 | Interleukin-10 signaling | 2.47 |
| 24 | HIV Transcription Initiation | 2.47 |
| 25 | Pausing and recovery of Tat-mediated HIV elongation | 2.46 |

**Depth 4.**

**Kept — active at depth 4**

| # | Pathway | $d_m$ |
|---|---|---|
| 1 | Intra-Golgi and retrograde Golgi-to-ER traffic | 3.97 |
| 2 | Glycerophospholipid biosynthesis | 3.96 |
| 3 | Post-translational modification: synthesis of GPI-anchored proteins | 3.95 |
| 4 | Golgi-to-ER retrograde transport | 3.95 |
| 5 | Factors involved in megakaryocyte development and platelet production | 3.95 |
| 6 | Semaphorin interactions | 3.95 |
| 7 | RUNX1 regulates genes involved in megakaryocyte differentiation and p... | 3.95 |
| 8 | FOXO-mediated transcription | 3.94 |
| 9 | Phospholipid metabolism | 3.94 |
| 10 | Kinesins | 3.94 |
| 11 | Chaperonin-mediated protein folding | 3.93 |
| 12 | Deubiquitination | 3.92 |
| 13 | Class I MHC mediated antigen processing & presentation | 3.92 |
| 14 | Signaling by Interleukins | 3.92 |
| 15 | SLC-mediated transmembrane transport | 3.90 |
| 16 | Coagulation pathway | 3.90 |
| 17 | Synthesis of PA | 3.90 |
| 18 | Regulation of clotting cascade | 3.90 |
| 19 | Regulation of MITF-M-dependent genes involved in apoptosis | 3.90 |
| 20 | MHC class II antigen presentation | 3.90 |
| 21 | Cell surface interactions at the vascular wall | 3.89 |
| 22 | Cellular Senescence | 3.88 |
| 23 | NOTCH3 Intracellular Domain Regulates Transcription | 3.88 |
| 24 | Ca2+ pathway | 3.87 |
| 25 | Mitotic Prometaphase | 3.85 |

**Erased — exit entering depth 4**

| # | Pathway | $d_m$ |
|---|---|---|
| 1 | Signaling by MET | 3.50 |
| 2 | Synthesis of substrates in N-glycan biosythesis | 3.50 |
| 3 | Signaling by NTRK2 (TRKB) | 3.49 |
| 4 | Tie2 Signaling | 3.49 |
| 5 | Depolymerization of the Nuclear Lamina | 3.49 |
| 6 | Complement cascade | 3.49 |
| 7 | COPI-mediated anterograde transport | 3.49 |
| 8 | HSF1-dependent transactivation | 3.49 |
| 9 | Differentiation of T cells | 3.49 |
| 10 | Amyloid fiber formation | 3.49 |
| 11 | Nuclear Events (kinase and transcription factor activation) | 3.49 |
| 12 | Respiratory Syncytial Virus Infection Pathway | 3.48 |
| 13 | Signaling by SCF-KIT | 3.48 |
| 14 | Translation of Replicase and Assembly of the Replication Transcriptio... | 3.48 |
| 15 | FLT3 Signaling | 3.48 |
| 16 | Regulation of MITF-M-dependent genes involved in lysosome biogenesis... | 3.48 |
| 17 | Mitochondrial tRNA aminoacylation | 3.48 |
| 18 | Triglyceride metabolism | 3.48 |
| 19 | Generation of second messenger molecules | 3.48 |
| 20 | Transcriptional regulation of granulopoiesis | 3.48 |
| 21 | CD28 dependent PI3K/Akt signaling | 3.48 |
| 22 | RNA Polymerase III Transcription Initiation | 3.48 |
| 23 | Metabolism of carbohydrates and carbohydrate derivatives | 3.47 |
| 24 | Heparan sulfate/heparin (HS-GAG) metabolism | 3.47 |
| 25 | Mitotic G2-G2/M phases | 3.47 |


\end{document}